\documentclass[10pt,journal,compsoc]{IEEEtran}
\usepackage[nocompress]{cite}
\usepackage[utf8]{inputenc}
\usepackage[T1]{fontenc}
\usepackage{hyperref}
\usepackage{url}
\usepackage{booktabs}
\usepackage{amsfonts}
\usepackage{nicefrac}
\usepackage{microtype}
\usepackage{xcolor}
\usepackage{amsmath,amsfonts,amsthm,amssymb}
\usepackage{graphicx}
\usepackage{subfigure}
\usepackage{enumitem}
\usepackage{multirow, bigdelim}
\usepackage{numprint}
\usepackage{pifont}
\usepackage{caption}

\newcommand{\cmark}{\ding{51}}
\newcommand{\header}[1]{{\vspace{+1mm}\flushleft \textbf{#1}}}

\begin{document}
\title{Amortized Auto-Tuning:\\Cost-Efficient Bayesian Transfer Optimization\\for Hyperparameter Recommendation}

\author{Yuxin Xiao, Eric P. Xing, \IEEEmembership{Fellow, IEEE,} and Willie Neiswanger
\IEEEcompsocitemizethanks{\IEEEcompsocthanksitem Yuxin Xiao is with Carnegie Mellon University; Eric P. Xing is with Carnegie Mellon University, Petuum, and MBZUAI; Willie Neiswanger is with Stanford University and Petuum.\protect}
\thanks{Manuscript received March, 2022.}}

\markboth{IEEE Transactions on Pattern Analysis and Machine Intelligence, ~Vol.~--, No.~--, Month~Year}
{Xiao \MakeLowercase{\textit{et al.}}: Amortized Auto-Tuning: Cost-Efficient Bayesian Transfer Optimization for Hyperparameter Recommendation}

\IEEEtitleabstractindextext{
\begin{abstract}
With the surge in the number of hyperparameters and training times of modern machine learning models, hyperparameter tuning is becoming increasingly expensive.
However, after assessing $40$ tuning methods systematically, we find that each faces certain limitations.
In particular, methods that speed up tuning via knowledge transfer typically require the final performance of hyperparameters and do not focus on low-fidelity information.
As we demonstrate empirically, this common practice is suboptimal and can incur an unnecessary use of resources.
It is more cost-efficient to instead leverage low-fidelity tuning observations to measure inter-task similarity and transfer knowledge from existing to new tasks accordingly.
However, performing multi-fidelity tuning comes with its own challenges in the transfer setting: the noise in additional observations and the need for performance forecasting.
Therefore, we propose and conduct a thorough analysis of a multi-task multi-fidelity Bayesian optimization framework, which leads to the best instantiation---\underline{A}mor\underline{T}ized \underline{A}uto-\underline{T}uning (\texttt{AT2}).
We further present an offline-computed 27-task \underline{Hyper}parameter \underline{Rec}ommendation (\textrm{HyperRec}) database to serve the community.
Extensive experiments on \textrm{HyperRec} and other real-world databases illustrate the effectiveness of our \texttt{AT2} method.
\end{abstract}

}

\maketitle
\IEEEdisplaynontitleabstractindextext
\IEEEpeerreviewmaketitle

\section{Introduction} \label{sec:1}

Modern machine learning models typically come with a large number of hyperparameters and are often sensitive to their values. 
Consequently, researchers have paid increasing attention to automatic hyperparameter tuning \cite{he2021automl, hutter2019automated, yao2018taking}, which aims to identify a set of optimal hyperparameters for a learning task without human experts. 
With the aid of optimization histories of past tuning sessions, some methods propose to accelerate new tuning processes via knowledge transfer.
Despite their impressive results, these methods come with limitations on their cost-efficiency and flexibility, in the sense that they either make modality-specific one-time predictions \cite{achille2019task2vec, cui2018large, Mittal_2020_CVPR, xue2019transferable, yang2019oboe} or rely on extra information from new tasks \cite{feurer2018scalable, jomaa2021hyperparameter, law2018hyperparameter, perrone2018scalable, salinas2020quantile}. 
More importantly, they generally operate on the final performance of hyperparameters and ignore low-fidelity information \cite{fusi2018probabilistic, snoek2015scalable, springenberg2016bayesian, swersky2013multi, wistuba2021few}.
As we demonstrate via a motivating example in Section~\ref{sec:2.3}, this practice incurs an unnecessary cost. 
It is more resource-efficient to instead utilize cheap-to-obtain low-fidelity tuning observations when carrying out inter-task hyperparameter transfer learning.

However, performing multi-fidelity tuning in the transfer setting is non-trivial.
It requires carefully distilling relevant knowledge from the additional multi-fidelity information in existing tasks.
It also demands accurate forecasting to extrapolate max-fidelity performance based on corresponding low-fidelity observations.
To this end, we resort to the well-established approach of Bayesian optimization (BO) and conduct a thorough analysis of a multi-task multi-fidelity BO framework.
More precisely, we address the aforementioned challenges by considering a family of kernels and, based on an extensive empirical evaluation, develop an \underline{A}mor\underline{T}ized \underline{A}uto-\underline{T}uning (\texttt{AT2}) method---the name stems from the fact that future tuning sessions will write off past tuning costs.

We summarize our contributions as follows:
(1) To examine the cost-efficiency and flexiblity of existing baselines, we study $40$ hyperparameter optimization methods based on seven specific criteria and demonstrate their limitations empirically.
(2) Inspired by this study, we aim to better leverage cheap-to-obtain low-fidelity observations for measuring inter-task dependency efficiently. In particular, we conduct a thorough analysis of the multi-task multi-fidelity BO framework where we empirically evaluate $64$ different instantiations.
(3) To motivate our analysis and as a service to the community, we present the \underline{Hyper}parameter \underline{Rec}ommendation (\textrm{HyperRec}) database. It consists of $27$ unique computer vision tuning tasks with $150$ distinct configurations over a $16$-dimensional hyperparameter space.
(4)~Based on the analysis, we propose \texttt{AT2}, a multi-task multi-fidelity BO method, which uses a novel task kernel and acquisition function. It outperforms competitive baselines on \textrm{HyperRec} and other real-world tuning databases.\footnote{\texttt{AT2} method and \textrm{HyperRec} database are available at \url{https://github.com/xiaoyuxin1002/amortized-auto-tuning}.}
\begin{table*}[h!]
  \centering
  \begin{tabular}{c|c|cccc|ccc} 
    \toprule
        \multirow{2}{*}{\textbf{Category}} & \multirow{2}{*}{\textbf{Method}} & \multicolumn{4}{c|}{\textbf{Cost-Efficiency}} & \multicolumn{3}{c}{\textbf{Flexibility}} \\
        & & \textbf{E1} & \textbf{E2} & \textbf{E3} & \textbf{E4} & \textbf{F1} & \textbf{F2} & \textbf{F3} \\
    \midrule 
    \midrule
        \multirow{3}{*}{\shortstack{Single-task\\ single-fidelity BO}} & \texttt{DNGO} \cite{snoek2015scalable} &  &  &  & \cmark & \cmark & \cmark & \\
        & \texttt{GPBO} \cite{snoek2012practical} &  &  & \cmark & \cmark & \cmark & \cmark &  \\
        & \texttt{ROAR} \cite{hutter2011sequential} &  &  & \cmark & \cmark & \cmark & \cmark &  \\
    \midrule
        \multirow{9}{*}{\shortstack{Multi-fidelity\\ BO}} & \texttt{Fabolas} \cite{klein2017fast} & & \cmark & \cmark & \cmark & \cmark & \cmark & \\
        & \texttt{TSE} \cite{hu2019multi} &  & \cmark & \cmark & \cmark & \cmark & \cmark & \\
        & \texttt{MF-GP-UCB} \cite{kandasamy2016gaussian} &  & \cmark & \cmark & \cmark & \cmark & \cmark & \\ 
        & \texttt{BOCA} \cite{kandasamy2017multi} & & \cmark & \cmark & \cmark & \cmark & \cmark & \\
        & \texttt{BOIL} \cite{nguyen2019bayesian} & & \cmark & \cmark & \cmark & \cmark & \cmark & \\
        & \texttt{MF-MES} \cite{takeno2020multi} & & \cmark & \cmark & \cmark & \cmark & \cmark & \\
        & \texttt{MF-PES} \cite{zhang2017information} & & \cmark & \cmark & \cmark & \cmark & \cmark & \\
        & \texttt{DNN-MFBO} \cite{li2020multi} & & \cmark & \cmark & \cmark & \cmark & \cmark & \\
        & \texttt{taKG} \cite{wu2020practical} & & \cmark & \cmark & \cmark & \cmark & \cmark & \\
    \midrule
        \multirow{3}{*}{\shortstack{Learning curve\\ modeling}} & \texttt{Freeze-Thaw} \cite{swersky2014freeze} &  & \cmark & \cmark & \cmark & \cmark & \cmark & \\
        & \texttt{LC Pred} \cite{domhan2015speeding} &  & \cmark & \cmark & \cmark & \cmark & \cmark & \\ 
         & \texttt{BO-BOS} \cite{dai2019bayesian} &  & \cmark & \cmark & \cmark & \cmark & \cmark & \\ 
    \midrule
        \multirow{4}{*}{\shortstack{Bandit-based\\ approach}} & \texttt{Hyperband} \cite{li2017hyperband} & & \cmark & \cmark & \cmark & \cmark & \cmark & \cmark \\
        & \texttt{BOHB} \cite{falkner2018bohb} &  & \cmark & \cmark & \cmark & \cmark & \cmark & \cmark \\
        & \texttt{MFES-HB} \cite{li2020mfes} &  & \cmark & \cmark & \cmark & \cmark & \cmark & \cmark \\
        & \texttt{ABLR-HB} \cite{valkov2018simple} & \cmark & \cmark & \cmark & \cmark & \cmark & \cmark & \cmark \\
    \midrule 
        \multirow{9}{*}{\shortstack{Multi-task\\ BO}} & \texttt{MTBO} \cite{swersky2013multi} & \cmark &  & \cmark & \cmark & \cmark & \cmark & \cmark \\
        & \texttt{BOHAMIANN} \cite{springenberg2016bayesian} & \cmark &  &  & \cmark & \cmark & \cmark & \cmark \\
        & \texttt{ABLR} \cite{perrone2018scalable} & \cmark &  &  & \cmark & \cmark & \cmark &  \\
        & \texttt{GCP} \cite{salinas2020quantile} & \cmark &  &  & \cmark & \cmark & \cmark & \\
        & \texttt{RGPE} \cite{feurer2018scalable} & \cmark &  &  & \cmark & \cmark & \cmark & \\
        & \texttt{FSBO} \cite{wistuba2021few} & \cmark & & & \cmark & \cmark & \cmark & \\
        & \texttt{Policy Search} \cite{letham2019bayesian} & \cmark &  &  & \cmark & \cmark & \cmark & \\
        & \texttt{DMFBS} \cite{jomaa2021hyperparameter} & \cmark &  &  & \cmark & \cmark &  & \cmark \\
        & \texttt{distGP} \cite{law2018hyperparameter} & \cmark &  &  & \cmark &  &  & \cmark \\
    \midrule
        \multirow{4}{*}{\shortstack{Warm-starting\\ method}} & \texttt{Siamese-BHO} \cite{kim2017learning} & \cmark &  &  & \cmark & \cmark &  & \cmark \\
        & \texttt{MI-SMBO} \cite{feurer2014using} & \cmark & & & \cmark & \cmark & & \cmark \\
        & \texttt{wsKG} \cite{poloczek2016warm} & \cmark &  &  & \cmark & \cmark & \cmark & \\
        & \texttt{Box BO} \cite{perrone2019learning} & \cmark &  &  & \cmark & \cmark & \cmark &  \\
    \midrule
        \multirow{4}{*}{\shortstack{Recommendation\\ method}} & \texttt{SCoT} \cite{bardenet2013collaborative} & \cmark &  &  & \cmark & \cmark & \cmark &  \\
        & \texttt{PMF} \cite{fusi2018probabilistic} & \cmark &  &  & \cmark & \cmark & \cmark & \\
        & \texttt{Data Grouping} \cite{xue2019transferable} & \cmark &  &  &  &  &  & \cmark \\
        & \texttt{OBOE} \cite{yang2019oboe} & \cmark &  &  &  &  &  & \cmark \\
    \midrule
        \multirow{4}{*}{\shortstack{Domain-specific\\ method}} & \texttt{task2vec} \cite{achille2019task2vec} & \cmark &  &  &  &  &  & \cmark \\
        & \texttt{DSTL} \cite{cui2018large} & \cmark &  &  &  &  &  & \cmark \\
        & \texttt{HyperSTAR} \cite{Mittal_2020_CVPR} & \cmark &  &  &  &  &  & \cmark \\
        & \texttt{TNP} \cite{wei2019transferable} & \cmark &  &  & \cmark &  &  &  \\
    \midrule 
        \multirow{2}{*}{\shortstack{Multi-task\\ multi-fidelity BO}} & \multirow{2}{*}{\texttt{AT2} [Ours]} & \multirow{2}{*}{\cmark} & \multirow{2}{*}{\cmark} & \multirow{2}{*}{\cmark} & \multirow{2}{*}{\cmark} & \multirow{2}{*}{\cmark} & \multirow{2}{*}{\cmark} & \multirow{2}{*}{\cmark} \\
        & & & & & & & & \\
    \bottomrule 
  \end{tabular}
  \caption{Cost-efficiency (\textbf{E}) and flexibility (\textbf{F}) of hyperparameter tuning methods. \textbf{E1}: transferable; \textbf{E2}: low-fidelity; \textbf{E3}: cost-aware; \textbf{E4}: sequential; \textbf{F1}: modality-agnostic; \textbf{F2}: self-contained; \textbf{F3}: cold-start friendly. Section~\ref{sec:2.2} explains the details.}
  \label{tab:full_existing} 
\end{table*}


\section{Background \& Motivation} \label{sec:2}

\subsection{Preliminaries} \label{sec:2.1}

Consider a black-box function $f:\mathcal{X} \rightarrow \mathbb{R}$ where the input space $\mathcal{X}$ is defined as the Cartesian product of a task space $\mathcal{T}$, a configuration space $\mathcal{C}$, and a fidelity space $\mathcal{E}$, i.e., $\mathcal{X} = \mathcal{T} \times \mathcal{C} \times \mathcal{E}$. 
The fidelity space can only be queried in incremental order. 
That is, at each iteration, we can make a (typically expensive) function evaluation and obatin a noisy observation $y = f(x) + \epsilon$ (where $\epsilon$ is drawn from some noise distribution) for an input $x = (t,c,e) \in \mathcal{X}$, only if we have already queried $(t,c,e')$ for all $e' < e$.
Consequently, to acquire an observation for a higher fidelity value, we incur a larger computational cost in terms of more query iterations.

Suppose for a set of \textit{past} tasks $\{t_i\}_{i=1}^T$, we have collected some subset of associated observations $\{y_i\}_{i=1}^N$ via querying the input space.
Given a \textit{new} task $t^*$, we would like to propose an optimization strategy that aims to identify $x^* = (t^*, c^*, e^*)$ using as little computation as possible with the help of knowledge transfer from past observations. 
Here, $c^*$ and $e^*$ are where $f$ achieves its maximum on $t^*$, i.e., $(c^*, e^*) = \arg\max_{c\in\mathcal{C},\, e\in\mathcal{E}} f((t^*, c, e))$.

In this paper, we focus on this setup for hyperparameter transfer optimization, where $t\in\mathcal{T}$ is a single tuning instance, $c\in\mathcal{C}$ is a hyperparameter configuration, $e\in\mathcal{E}$ is an epoch value, and $f(x)$ is the associated validation accuracy.
Following \cite{yang2019oboe, li2017hyperband}, we treat $\mathcal{C}$ as a finite discrete set of pre-selected configurations.
On a new task $t^*$, we evaluate an optimization strategy via two metrics: given iteration budget $Q$, we want to minimize the \textit{simple regret} $R_Q = f(x^*) - \max_{q=1,\dots,Q} f(x_q)$ of queried points $x_q = (t^*, c_q, e_q)$, and maximize the \textit{final performance} $F_Q = \max_{e\in\mathcal{E}} f((t^*, c_Q^*, e))$ of the predicted optimal configuration $c^*_Q$.

\begin{figure*}[t!]
    \centering
    \includegraphics[width=\textwidth]{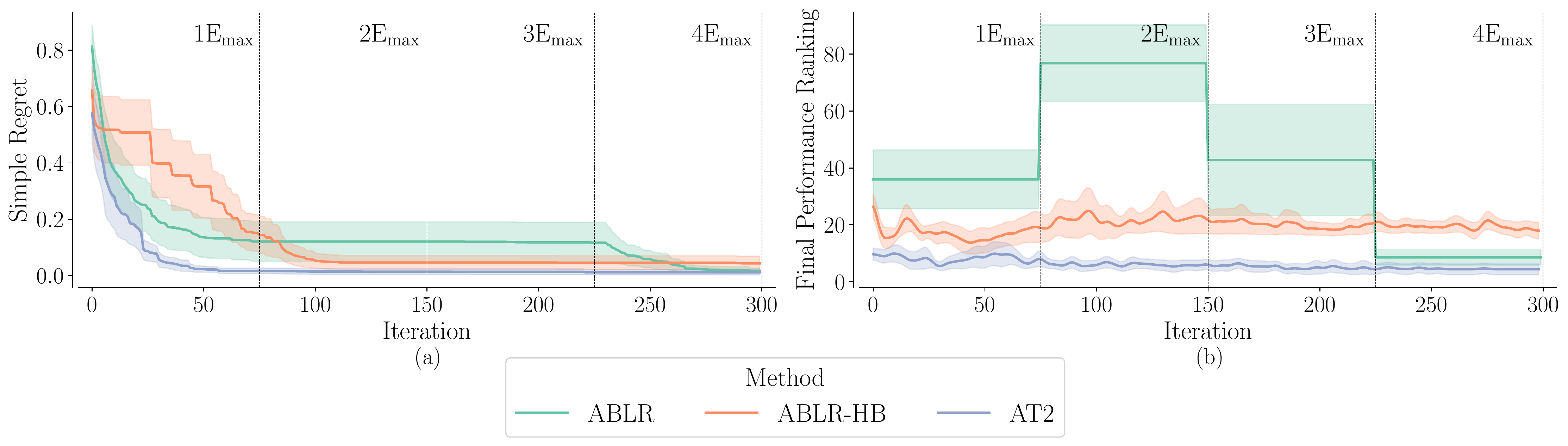}
    \caption{An example illustrating the advantages of leveraging low-fidelity observations for cost-efficient hyperparameter transfer optimization, shown on the \textrm{HyperRec} database (Section~\ref{sec:3.1}).
    Each method is allowed an iteration budget four times the max-fidelity ($\text{E}_{\max}$). The results are averaged across five train-test task pairs. Lower is better. The shaded regions represent one standard error of each method. Our proposed \texttt{AT2} method conducts knowledge transfer based on cheap-to-get low-fidelity information and consistently outperforms the other baselines in both metrics.}
    \label{fig:motivation}
\end{figure*}

\subsection{Limitations of Existing Work} \label{sec:2.2}

Besides the performance metrics introduced above, we further assess the cost-efficiency (\textbf{E}) and flexibility (\textbf{F}) of hyperparameter tuning methods with the following criteria.

\begin{description}
    \item [\textbf{E1}:] (\textit{Transferable}) The method should leverage the observations of existing tasks and perform knowledge transfer to speed up the tuning process of any new tasks.
    \item [\textbf{E2}:] (\textit{Low-fidelity}) The method should utilize low-fidelity information instead of just max-fidelity performance.
    \item [\textbf{E3}:] (\textit{Cost-aware}) The method should respond to different computational costs involved in querying for observations of different tasks or fidelities.
    \item [\textbf{E4}:] (\textit{Sequential}) The method should actively adapt to any observations received while tuning new tasks and carry out feedback-driven sequential tuning.
    \item [\textbf{F1}:] (\textit{Modality-agnostic}) The method should be modality-free and broadly applicable to various data types.
    \item [\textbf{F2}:] (\textit{Self-contained}) The method should operate on any new tasks without relying on auxiliary information such as extracted metadata or pre-computed representations.
    \item [\textbf{F3}:] (\textit{Cold-start friendly}) The method should work well under cold start situations and not require any observations on new tasks in order to execute.
\end{description}

To this end, we summarize results of an evaluation on $40$ existing hyperparameter recommendation methods in Table~\ref{tab:full_existing}.
In general, widely-adopted Bayesian optimization (BO) methods \cite{mockus1978application} define a surrogate model
(e.g., Gaussian process \cite{kandasamy2020tuning, snoek2012practical}, neural networks \cite{snoek2015scalable}, or random forests \cite{hutter2011sequential}) on the target black-box function and deploy an acquisition function (e.g., GP-UCB \cite{auer2002using, srinivas2010gaussian} or GP-EI \cite{jones1998efficient}) to determine future query points. However, vanilla BO methods focus on single-task single-fidelity tuning and require preliminary observations for proper initialization.

One line of extension is multi-fidelity BO methods \cite{hu2019multi, kandasamy2016gaussian, klein2017fast}, which apply cheap approximations to the target function. 
Some bandit-based \cite{falkner2018bohb, li2017hyperband, li2020mfes} and learning curve modeling \cite{domhan2015speeding, swersky2014freeze} approaches also examine the multi-fidelity information for early stopping.
Hence, they do a better job at leveraging low-fidelity information for single-task tuning and allocate resources more effectively.

In another direction, multi-task BO methods aim to transfer knowledge between multiple optimization problems via methods such as multi-output GPs \cite{feurer2018scalable, swersky2013multi}, Bayesian deep learning \cite{springenberg2016bayesian, wistuba2021few}, and Bayesian linear regression \cite{perrone2018scalable}.
However, these methods typically operate on the final (max-fidelity) performance and treat all queries as having equal cost.
Even when equipped with the optimization histories of previous tasks, these methods often spend some budget on obtaining initial max-fidelity observations on new tasks so as to measure inter-task similarity before carrying out knowledge transfer.
As we will show in the example below, this procedure is unnecessary and, often, unexpectedly costly.

Towards a similar goal, some approaches view hyperparameter transfer learning from a warm-start \cite{feurer2014using, kim2017learning, perrone2019learning, poloczek2016warm} or recommendation \cite{bardenet2013collaborative, fusi2018probabilistic, xue2019transferable, yang2019oboe} perspective and typically rely on pre-computed task-dependent metadata or representations.
A few prediction-only methods \cite{achille2019task2vec, cui2018large, Mittal_2020_CVPR, wei2019transferable} have been proposed for specific domains but satisfy neither the sequential tuning nor modality-agnostic criteria.

On account of the analysis above, we draw merits from both multi-task and multi-fidelity BO methods and present an \underline{A}mor\underline{T}ized \underline{A}uto-\underline{T}uning (\texttt{AT2}) method, which fulfills each of the cost-efficiency and flexibility criteria. 
In particular, \texttt{AT2} performs sequential, modality-free tuning of validation accuracy and transfers knowledge from existing to new tasks effectively, even under cold start scenarios, based on cheap-to-obtain low-fidelity observations (instead of auxiliary or full-fidelity information). 
Nevertheless, it is a non-trivial task to consider the multi-fidelity information in the transfer setting. For instance, the additional structure given by the multi-fidelity observations from previous tasks demands careful attention. Additionally, in this setting we must forecast the max-fidelity performance of hyperparameters based on their corresponding low-fidelity observations.

\subsection{Motivating Example} \label{sec:2.3}

To illustrate the challenges involved in multi-task multi-fidelity tunning, we implement a motivating example where \texttt{AT2} is compared against two multi-task baselines under the cold start situation: \texttt{ABLR} \cite{perrone2018scalable}, which ignores multi-fidelity information, and \texttt{ABLR-HB} \cite{valkov2018simple}, which processes multi-fidelity information via the \texttt{Hyperband} \cite{li2017hyperband} regime.
Here, we utilize a real-world hyperparameter tuning database, \textrm{HyperRec} (which will be introduced in detail in Section~\ref{sec:3.1}), and report the results averaged over five train-test task pairs based on the two metrics ($R_Q$, $F_Q$) discussed in Section~\ref{sec:2.1}.
For $F_Q$, we use the \textit{final performance ranking} of the configuration with the highest predicted mean instead of the raw score for a clearer presentation. All three methods are trained with their respective default settings and given an iteration budget four times the max fidelity (i.e., $Q = 4 \times \text{E}_{\max} = 300$).

As shown in Figure~\ref{fig:motivation}, since \texttt{ABLR} uses max-fidelity observations from the test task, its final performance ranking suffers initially and improves only after making several max-fidelity queries.
Meanwhile, when \texttt{ABLR} is still waiting for its first max-fidelity feedback, \texttt{AT2} and \texttt{ABLR-HB} are able to update their predictions immediately after receiving low-fidelity feedback and quickly recognize promising hyperparameter configurations. This phenomenon renders the cost of max-fidelity initialization in \texttt{ABLR} unnecessary and illustrates the advantage of low-fidelity tuning.

On the other hand, although \texttt{Hyperband} is ideal for the parallel tuning setting when substantial computational resources are accessible, it begins by selecting a large batch of configurations and thus uses excessive computation in the low-fidelity region given the same iteration budget (in terms of total computation, disregarding parallelism).
As a result, in our sequential setting, \texttt{ABLR-HB} only achieves a lower simple regret than \texttt{ABLR} at around the $80$th iteration.
Moreover, \texttt{ABLR-HB}'s predicted final performance ranking declines after the $50$th iteration.
Since \texttt{ABLR-HB} treats the fidelity as a contextual variable, when more multi-fidelity observations become available on the test task, the extrapolation performance begins to suffer, as it becomes difficult to identify an informative subset of training observations and forecast full-fidelity performance.
Eventually, \texttt{ABLR} outperforms \texttt{ABLR-HB} in both metrics after initializing on enough full-fidelity observations.

Unlike \texttt{ABLR-HB}, which leaves the task of multi-fidelity tuning to \texttt{Hyperband}, our proposed \texttt{AT2} method sequentially selects queries for increased cost-efficiency. It quantifies inter-task dependencies based on low-fidelity information and converges to a high-ranking configuration thanks to careful forecasting of validation accuracies. In addition, \texttt{AT2} also balances exploration and exploitation well and achieves a lower simple regret than \texttt{ABLR} from the beginning. To this end, we focus on developing a multi-task multi-fidelity BO framework and discuss \texttt{AT2} in detail, in the next section.
\section{Methods} \label{sec:3}

\subsection{\textrm{HyperRec} Database} \label{sec:3.1}

We illustrate the problem setting in Section~\ref{sec:2.1} with an offline-computed \underline{Hyper}parameter \underline{Rec}ommendation database---\textrm{HyperRec}. 
\textrm{HyperRec} consists of $27$ unique image classification tuning tasks, each with $150$ distinct configurations composed of $16$ nested hyperparameters. Each task is evaluated on each configuration for $75$ epochs and repeated with two different seeds. We record the validation loss and top one, five, and ten accuracies in \textrm{HyperRec}. 

To the best of our knowledge, this is the first hyperparameter recommendation database specifically targeting computer vision tasks.
Appendix~\ref{app:1} explains \textrm{HyperRec} and compares it with other related databases in detail.
By releasing \textrm{HyperRec}, we seek to serve both the hyperparameter tuning and computer vision communities with a database for testing and comparing the performance of existing and future hyperparameter tuning or image classification algorithms.

\subsection{Multi-Task Multi-Fidelity BO Framework} \label{sec:3.2}

In what follows, we will describe the overall multi-task multi-fidelity BO framework and then give an extensive study of different implementations of this framework.
The BO paradigm is characterized by the use of a probabilistic surrogate model of the expensive black-box target $f(x)$.
In this paper, we stick to the popular choice of Gaussian process (GP) for the surrogate model due to its accurate uncertainty quantification.
To enhance the model scalability, we adopt the stochastic variational GP regression framework \cite{hensman2015scalable}.

A GP over the input space $\mathcal{X}$ is a random process from $\mathcal{X}$ to $\mathbb{R}$, represented by a mean function $\mu: \mathcal{X} \rightarrow \mathbb{R}$ and a kernel (i.e., covariance function) $\kappa: \mathcal{X}^2 \rightarrow \mathbb{R}_+$. If $f \sim \mathcal{GP}(\mu, \kappa)$, then we have $f(x) \sim \mathcal{N}(\mu(x), \kappa(x,x))$ for all $x \in \mathcal{X}$. Consider $N$ collected observations $D_N = \{(x_i, y_i)\}_{i=1}^N$ from $T$ tasks where $y_i = f(x_i) + \epsilon_i \in \mathbb{R}$ and $\epsilon_i \sim \mathcal{N}(0, \eta^2)$.
We stack $D_N$ to form $\mathbf{X} \in \mathcal{X}^N$ and $\mathbf{Y} \in \mathbb{R}^N$. When using variational inference, we also learn $M$ inducing inputs $\mathbf{Z} \in \mathcal{X}^M$ where $M \ll N$ and the corresponding inducing variables $\mathbf{u} = f(\mathbf{Z})$.
Here, we let the prior distribution $p(\mathbf{u}) = \mathcal{N}(\mathbf{0}, \mathbf{I})$ and the variational distribution $q(\mathbf{u}) = \mathcal{N}(\mathbf{m}, \mathbf{S})$.
We optimize the GP hyperparameters $\theta$ and variational parameters $\varphi$ by maximizing the variational evidence lower bound (ELBO):
\begin{align} \begin{aligned} \label{eq:elbo}
    \theta^*, \varphi^* 
        = \; & \underset{\theta,\,\varphi}{\mathrm{argmax}} \, \sum_{i=1}^N \mathbb{E}_{q_{\theta,\varphi}(f(x_i))} \left[\log p_\theta(y_i | f(x_i)) \right] \\ 
        & - \text{KL} \left[ q_\varphi(\mathbf{u}) \| p(\mathbf{u}) \right],
\end{aligned} \end{align}
where $q_{\theta,\varphi}(f(x_i))$ is the marginal of $p_\theta(f(x_i)|\mathbf{u}) q_\varphi(\mathbf{u})$. The predictive distribution for query $x$ is
\begin{align} \begin{aligned} \label{eq:gp_posterior}
    p_\theta(f(x) | D_N)  
        & \approx \int p_\theta(f(x) | \mathbf{u}) q_\varphi(\mathbf{u}) \, \mathrm{d}\mathbf{u} \\
        & = \mathcal{N} ( \mathbf{A}\mathbf{m}, \, \kappa(x,x) +  \mathbf{A} (\mathbf{S}-\mathbf{K}) \mathbf{A}^\top ),
\end{aligned} \end{align}
where $\mathbf{A} = \mathbf{k}\mathbf{K}^{-1}$, $\mathbf{k}\in\mathbb{R}^{1 \times M}$ with $\mathbf{k}_i = \kappa (x, z_i)$, and $\mathbf{K} \in \mathbb{R}^{M \times M}$ with $\mathbf{K}_{i,j} = \kappa (z_i, z_j)$.

After incorporating the information from previous tasks, we can construct an acquisition function $\phi: \mathcal{X} \rightarrow \mathbb{R}$ for the new tuning task $t^*$.
At iteration $q$, the next point to query, $x_q =  (t^*, c_q, e_q)$, is determined by maximizing the acquisition function to choose a configuration $c_q$ and running one
additional epoch (more details in Section~\ref{sec:3.4}).
After querying $f(x_q) = f((t^*, c_q, e_q))$, we collect the observation $y_q$, and then update the model parameters according to Equation~\ref{eq:elbo}.
This iterative process continues until we spend the iteration budget $Q$.

Next, we provide a thorough analysis of this general multi-task multi-fidelity BO framework, where we focus on a comparison of options for the key component---the kernel $\kappa$ in Section~\ref{sec:3.3}---and conclude with the best instantiation---\underline{A}mor\underline{T}ized \underline{A}uto-\underline{T}uning (\texttt{AT2}) algorithm in Section~\ref{sec:3.4}.

\begin{figure*}[t!]
    \centering
    \includegraphics[width=\textwidth]{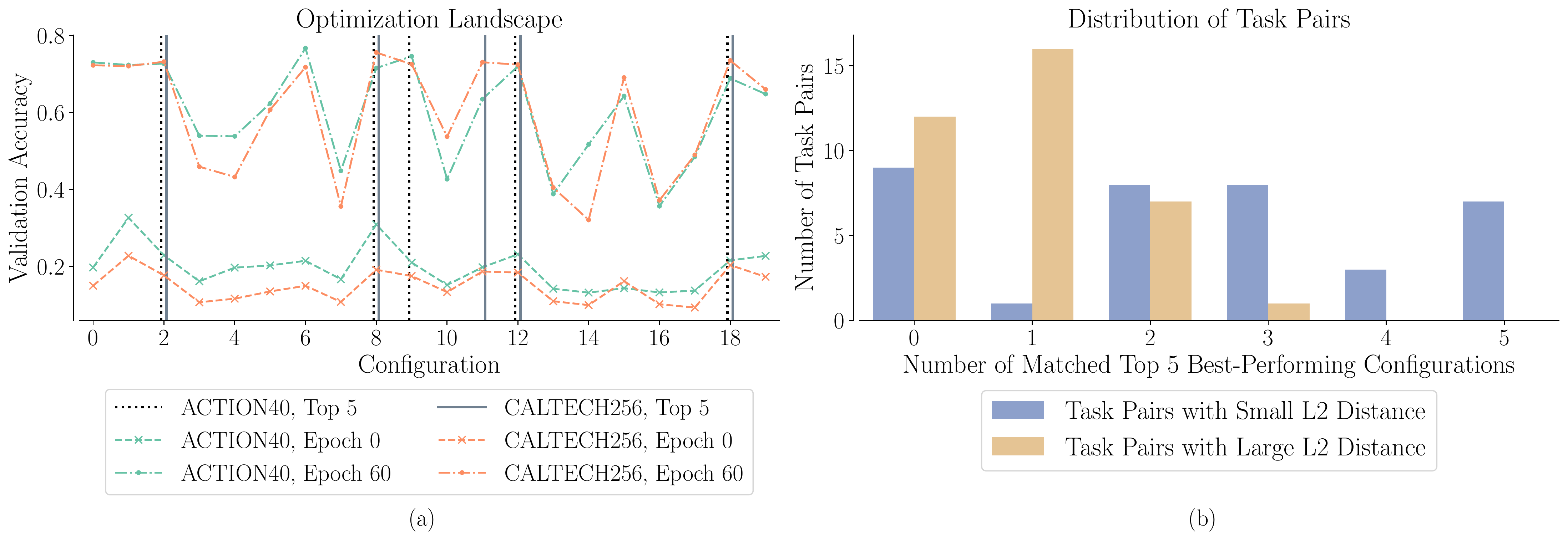}
    \caption{\textbf{(a)} Optimization landscapes of two tasks. \textrm{ACTION40} and \textrm{CALTECH256} have similar optimization landscapes and share a large portion of the top five best-performing configurations. \textbf{(b)} Relationship between inter-task L2 distance and the number of matched high-performing configurations. The figure shows the distributions of the top $10\%$ of task pairs whose zero-meaned optimization landscapes are closest or farthest in terms of L2 distance. Task pairs with smaller L2 distances are more likely to share high-performing configurations.}
    \label{fig:landscape}
\end{figure*}

\subsection{Kernel Analysis} \label{sec:3.3}

Since the input space $\mathcal{X}$ is the product of three spaces, we can use a kernel with the structure
\begin{align} \begin{aligned} \label{eq:kernel_expansion}
    \kappa(x, x') 
        & = \kappa((t,c,e), (t',c',e')) \\ 
        & = \kappa_\mathcal{T}(t, t') \otimes \kappa_\mathcal{C}(c, c') \otimes \kappa_\mathcal{E}(e, e'),
\end{aligned} \end{align}
where $\otimes$ denotes the Kronecker product. $\kappa_\mathcal{T}: \mathcal{T}^2 \rightarrow \mathbb{R}_+$, $\kappa_\mathcal{C}: \mathcal{C}^2 \rightarrow \mathbb{R}_+$, and $\kappa_\mathcal{E}: \mathcal{E}^2 \rightarrow \mathbb{R}_+$ are the task, configuration, and fidelity kernels, respectively. Below we will discuss suitable kernels for each of the three spaces and then carry out an empirical evaluation to find the best combination.

\subsubsection{Task Kernel} \label{sec:3.3.1}

Based on the example in Section~\ref{sec:2.3}, we are motivated to take advantage of low-fidelity function queries to define inter-task similarity.
One key observation we derived from \textrm{HyperRec} is that if two tasks have similar low-fidelity behaviors, they are \textit{more likely} to share high-scoring configurations.
For instance, \textrm{ACTION40} \cite{yao2011human} and \textrm{CALTECH256} \cite{griffin2007caltech} are two tasks in \textrm{HyperRec}.
As shown in Figure~\ref{fig:landscape} (a), they exhibit similar low-fidelity behavior, and their top five best-performing configurations largely overlap.

Moreover, we can see that this observation is ubiquitous among all task pairs. We select two groups of pairs according to the L2 distance of their optimization landscapes (i.e., between their zero-meaned validation accuracies). As shown in Figure~\ref{fig:landscape} (b), task pairs similar in L2 distance share larger portions of high-performing configurations.
We, therefore, propose the \texttt{OptiLand} task kernel, which infers the similarity between a new tuning task and past tasks by comparing their low-fidelity performance, in order to perform efficient optimization via knowledge transfer.

Consider a task pair $(t_1, t_2)$ and their respective queries $X_{1} = \{(t_1, c_{1,i}, e_{1,i})\}_{i=1}^{N_1}$ and $X_{2} = \{(t_2, c_{2,i}, e_{2,i})\}_{i=1}^{N_2}$.
Based on the setup in Section~\ref{sec:2.1}, $\mathcal{C}$ is a finite discrete set of pre-selected configurations and $\mathcal{E}$ is a finite discrete set of epoch values.
Therefore, we focus on finding an overlapping subset of queries in $t_1$ and $t_2$ so as to measure their inter-task similarity.
As reflected in the motivating example in Section~\ref{sec:2.3}, this provides robustness against the noise in the multi-fidelity observations.
More specifically, we define a matching function to return the set of configuration-fidelity tuples for which we have queried for observations on both tasks: $M(t_1, t_2) = \{(c,e) \,|\, (t_1,c,e)\in X_1, (t_2,c,e)\in X_2\}$.
The corresponding observation vectors are $\mathbf{Y}_{1|(t_1, t_2)}, \mathbf{Y}_{2|(t_1, t_2)} \in \mathbb{R}^{|M(t_1, t_2)|}$, respectively. 
Entries in $\mathbf{Y}_{1|(t_1, t_2)}$ and $\mathbf{Y}_{2|(t_1, t_2)}$ are min-max normalized to $[0,1]$, shifted to have zero mean, and ordered by a common permutation of $M(t_1, t_2)$.

Hence, the distance function between the optimization landscapes of $t_1$ and $t_2$ is $D(t_1, t_2) = \frac{||\mathbf{Y}_{1|(t_1, t_2)} - \mathbf{Y}_{2|(t_1, t_2)}||_2^2}{|M(t_1, t_2)|}$.
Since the observation vectors are normalized, we have $D(t_1, t_2) \in [0, 1]$.
Note that, when the number of matched query pairs $|M(t_1, t_2)| \leq 1$ (e.g., during initialization), we simply
make a na\"ive guess by setting $D(t_1, t_2)$ as an \textit{average} distance of $\frac{1}{2}$.
Note that both observation vectors are transformed to have zero mean, since we are interested in whether the optimization landscapes of two tasks have similar shapes.
In this way, two tasks will have zero distance if one's landscape is equal to another's shifted or scaled.

We then define the \texttt{OptiLand} task kernel to assess the dependency between $t_1$ and $t_2$ as follows:
\begin{align} \begin{aligned} \label{eq:task_kernel}
    \kappa_\mathcal{T} (t_1, t_2) = \exp\left(\frac{- D(t_1, t_2)}{(\xi \cdot \gamma(t_1, t_2))^2}\right) ,
\end{aligned} \end{align}
where $\xi \in \mathbb{R}_+$ is the length scale and 
\begin{align} \begin{aligned} \label{eq:scale_func}
    \gamma(t_1, t_2) = \frac{U}{1 + (U-1) \cdot R(t_1, t_2)}. 
\end{aligned} \end{align}
Here, $R(t_1, t_2) = \frac{|M(t_1, t_2)|}{|\mathcal{C}| \times |\mathcal{E}|}$ is the ratio of matched queries. $\gamma: \mathcal{T}^2 \rightarrow \mathbb{R}_+$ is a scaling function indicating the amount of information we have about the task pair.
Intuitively, no matter how many observations we have on $t_1$ and $t_2$ separately, if we have very few \textit{matched} pairs of observations, we are less confident about how well $D(t_1, t_2)$ captures the true difference between their optimization landscapes.
In this case, we would like to bias the tuning process of the new task towards existing tasks and increase the length scale to allow more knowledge transfer.

\begin{table*}[t!]
  \centering
  \begin{tabular}{c|c|c|c|c|c} 
    \toprule
        \textbf{Rank} & \textbf{Figure} & \textbf{Task Kernel} & \textbf{Configuration Kernel} & \textbf{Fidelity Kernel} & \textbf{ELBO} \\
    \midrule
    \midrule
        1 & Figure~\ref{fig:analysis} (a) & \texttt{OptiLand} & \texttt{DeepPoly} & \texttt{AccCurve} & 1.4951 \\
        49 & Figure~\ref{fig:analysis} (b) & \texttt{DeepPoly} & \texttt{DeepPoly} & \texttt{Matern} & 0.3053 \\
        64 & Figure~\ref{fig:analysis} (c) & \texttt{MTBO} & \texttt{Tree} & \texttt{Fabolas} & 0.1342 \\
    \bottomrule
  \end{tabular}
  \caption{Quantitative performance of different kernel compositions. The combination of our proposed \texttt{OptiLand} task kernel, \texttt{DeepPoly} configuration kernel, and \texttt{AccCurve} fidelity kernel gives the highest ELBO score among the $64$ candidates. Full results in Appendix~\ref{app:2}.}
  \label{tab:analysis}
\end{table*}

\begin{figure*}[t!]
    \centering
    \includegraphics[width=0.87\textwidth]{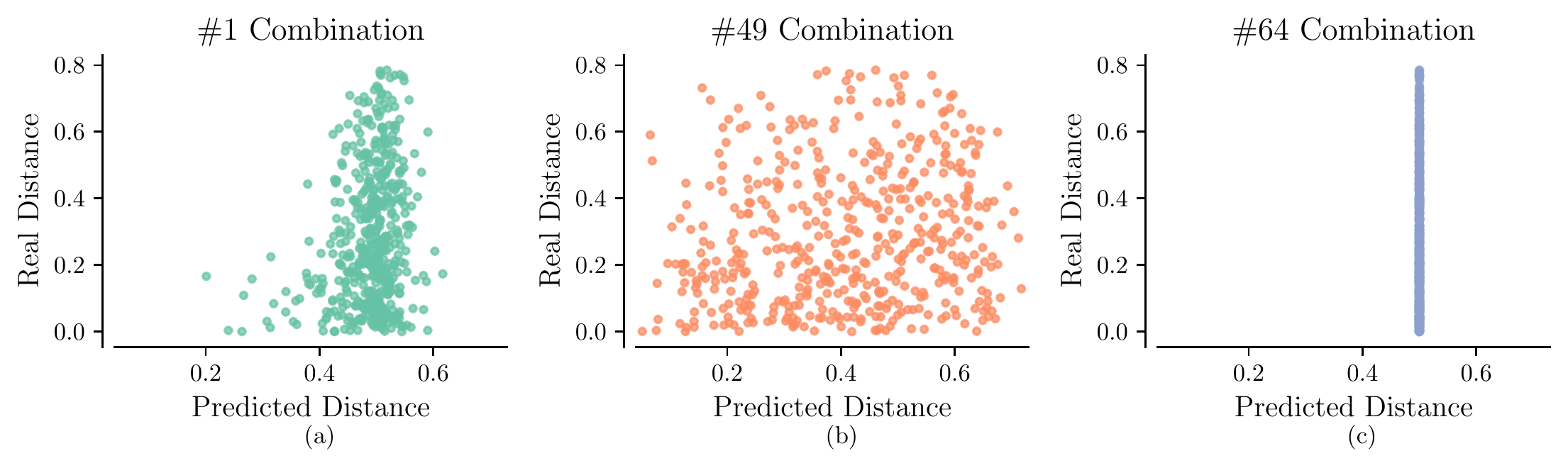}
    \caption{Qualitative performance of different kernel compositions. The highest-ranked kernel combination gives predicted distance that is relatively proportional to the real distance between query observations. Full results in Appendix~\ref{app:2}.}
    \label{fig:analysis}
\end{figure*}

Therefore, we use a learnable parameter $U$ to bound $\gamma \in [1, U]$.
When the ratio of matched queries $R(t_1, t_2) = 1$, then $\gamma=1$, and we leave the length scale alone and let the task kernel control the amount of information transfer. Alternatively, when $R(t_1, t_2) = 0$ and $\gamma=U$, we scale up the length scale by $U$ to increase the amount of information transfer.
This design is especially useful for the cold start situation. During the early search phase of the new task, configurations queried for the new and old tasks hardly overlap. 
Hence, we increase the length scale $\xi$ via $\gamma$ to allow more meaningful knowledge transfer for warm-starting the new task.
That is, we assume the new task is similar to existing tasks and try out configurations that perform well on existing tasks for the new task.
This artificial upscaling on $\xi$ is mitigated as more queries are made on the new task. 
Again, our claim is well supported by the good initial performance of \texttt{AT2} under the cold start scenario in the motivating example (Section~\ref{sec:2.3}).

Besides the proposed \texttt{OptiLand} task kernel, we compare three other possibilities for $\kappa_\mathcal{T}$.
\texttt{MTBO} task kernel \cite{swersky2013multi} is defined by a lookup table and optimized by learning the entries in the Cholesky decomposition of the covariance matrix. Some methods \cite{springenberg2016bayesian} also suggest learning an embedding for each task and apply a linear or second-order polynomial kernel on top of it. We name these two alternatives as \texttt{DeepLinear} and \texttt{DeepPoly} task kernels, respectively.

\subsubsection{Configuration Kernel} \label{sec:3.3.2}

The configuration kernel needs to deal with different hyperparameter types (i.e., numerical or categorical) and partially overlapping hyperparameter configurations. Therefore, \texttt{Tree} configuration kernel \cite{ma2020additive} advocates treating the configuration space as tree-structured and composites individual hyperparameter kernels in a sum-product way. For our experiments, we use an RBF or index kernel as the individual kernel for numerical or categorical hyperparameters, respectively. Alternatively, \texttt{Flat} configuration kernel discards the tree structure and multiplies all the individual kernels together.
Some prior work \cite{perrone2018scalable, springenberg2016bayesian} recommends encoding the configuration space via deep learning.
Hence, we construct a two-layer fully-connected neural network with \textit{tanh} activation function and learn an embedding input for each categorical hyperparameter.
Then \texttt{DeepLinear} or \texttt{DeepPoly} configuration kernels leverage a linear or second-order polynomial kernel based on network outputs, respectively.

\subsubsection{Fidelity Kernel} \label{sec:3.3.3}

The fidelity kernel aims to capture how the validation accuracy changes over epochs. As inspired by \cite{swersky2014freeze}, we define \texttt{AccCurve} fidelity kernel as a weighted integration over infinite basis functions: $\kappa_\mathcal{E} (e, e') = \int_0^\infty (1-\exp\{-\lambda e\}) (1-\exp\{-\lambda e'\}) \,\mathrm{d} \psi(\lambda)$. When the mixing measure $\psi$ takes the form of a gamma distribution with parameters $\alpha, \beta > 0$, the equation can be simplified into $\kappa_\mathcal{E} (e,e') = 1 + (\frac{\beta}{e+e'+\beta})^\alpha - (\frac{\beta}{e+\beta})^\alpha - (\frac{\beta}{e'+\beta})^\alpha$. Since the basis function approximates the shape of learning curves, this kernel extrapolates the high-fidelity performance well based on low-fidelity observations. Another popular kernel for modeling multi-fidelity information is \texttt{Fabolas} fidelity kernel \cite{klein2017fast}, where the authors assume a monotonic behavior of function evaluations with fidelity $e$. We also consider two simple choices for $\kappa_\mathcal{E}$---\texttt{RBF} and \texttt{Matern} fidelity kernels.

\begin{figure*}[t!]
    \centering
    \includegraphics[width=0.83\textwidth]{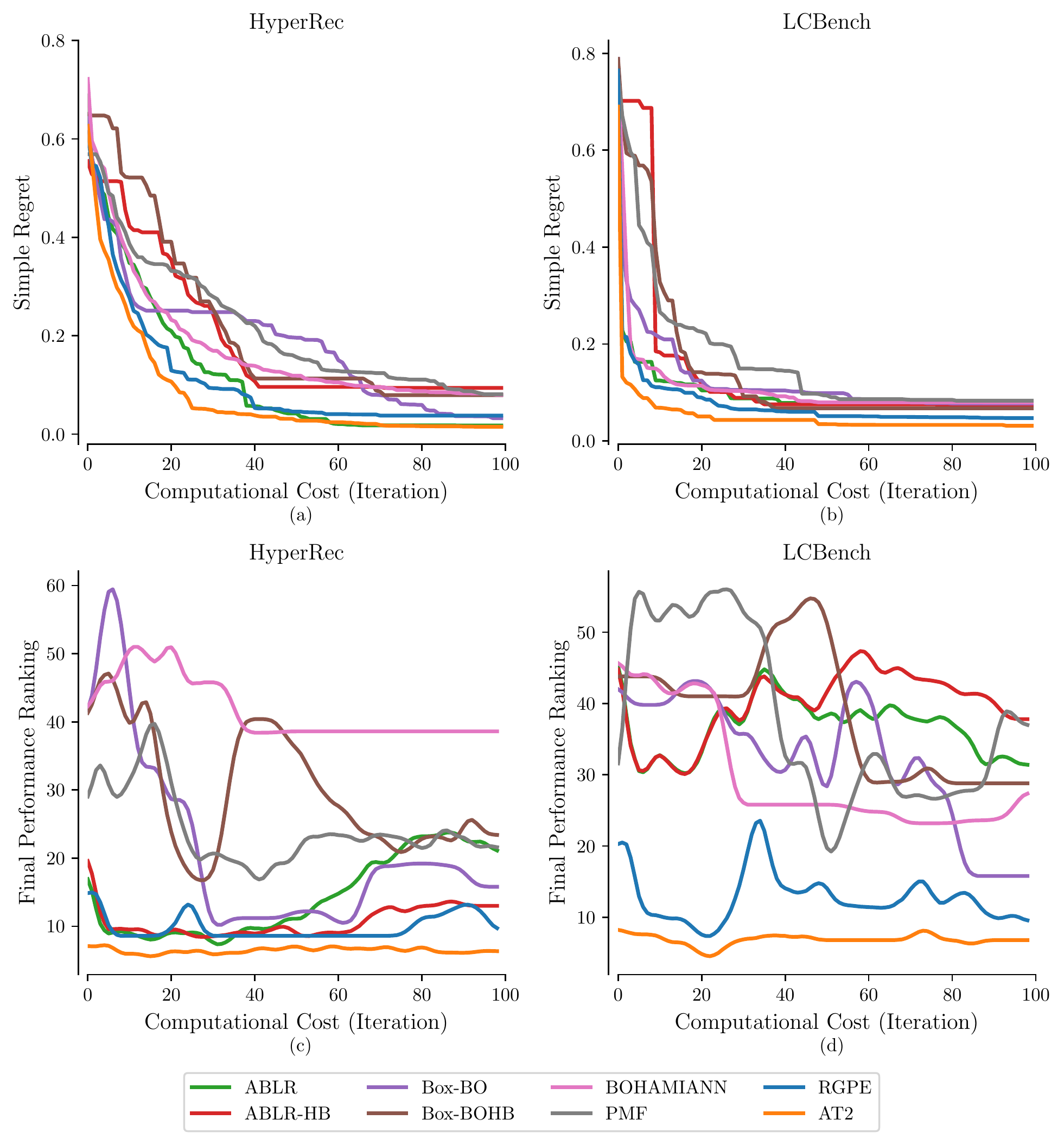}
    \caption{Performance of methods on \textrm{HyperRec} and \textrm{LCBench}. The results based on two metrics (simple regret and final performance ranking) are averaged across five train-test task pairs for each database. Lower is better. The predicted final performance rankings are smoothed with a hamming window of $10$ iterations.
    Our \texttt{AT2} method consistently achieves lower values in both metrics.
    We omit error bars here for readability and show a version with one standard error in Appendix~\ref{app:3}.}
    \label{fig:experiment}
\end{figure*}

\subsubsection{Empirical Evaluation} \label{sec:3.3.4}

In the previous sections, we discussed four options for each of the three component kernels, which gives rise to $64$ combinations.
Here, we assess the effectiveness of each option under our multi-task multi-fidelity BO framework and identify the best-performing instantiation based on an empirical evaluation.
More specifically, we randomly sample four tasks from \textrm{HyperRec}, which consists of \numprint{30000} observations.
We further sample \numprint{1000} observations from them to form the test set and use the rest as the train set. All of the kernel combinations are trained with the same setting introduced in Section~\ref{sec:3.2}.
Since we are concerned with how well different kernel combinations can explain the data, we report the ELBO on the test set as the quantitative metric in Table~\ref{tab:analysis} (full results in Appendix~\ref{app:2}).

In general, although the other three task kernels show competitive results, the \texttt{OptiLand} task kernel has the best performance with different configuration and fidelity kernels.
We attribute the success to its careful measurement of inter-task similarities based on matched queries.
For the configuration kernel, the neural network-based kernels (i.e., \texttt{DeepLinear} and \texttt{DeepPoly}) perform better than the other two options.
Finally, \texttt{AccCurve} fidelity kernel better models the shape of learning curves than the other three alternatives.
Among the $64$ candidates, the combination of \texttt{OptiLand} task kernel, \texttt{DeepPoly} configuration kernel, and \texttt{AccCurve} fidelity kernel achieves the highest ELBO value. Therefore, we will leverage this composition for our \texttt{AT2} method.

We also provide a qualitative analysis of the kernel performance in Figure~\ref{fig:analysis} (full results in Appendix~\ref{app:2}).
In particular, we divide the \numprint{1000} observations in the test set into $500$ pairs and compare their \textit{predicted distance}, defined as $(1 - \frac{\kappa(x,x')}{\sqrt{\kappa(x,x) \kappa(x',x')}})/2 \in [0,1]$ (given by the covariance kernel) with the \textit{true distance}, defined by the absolute difference in observations (which is bounded by $[0,1]$).
As shown in Figures~\ref{fig:analysis} (b) and (c), an ineffective kernel combination either give random or equal predicted distance regardless of the true distance.
In contrast, the predicted distance by the best-performing kernel combination is \textit{more aligned} with the true distance, which helps justify its quantitative result.

\begin{figure*}[h!]
    \centering
    \includegraphics[width=\textwidth]{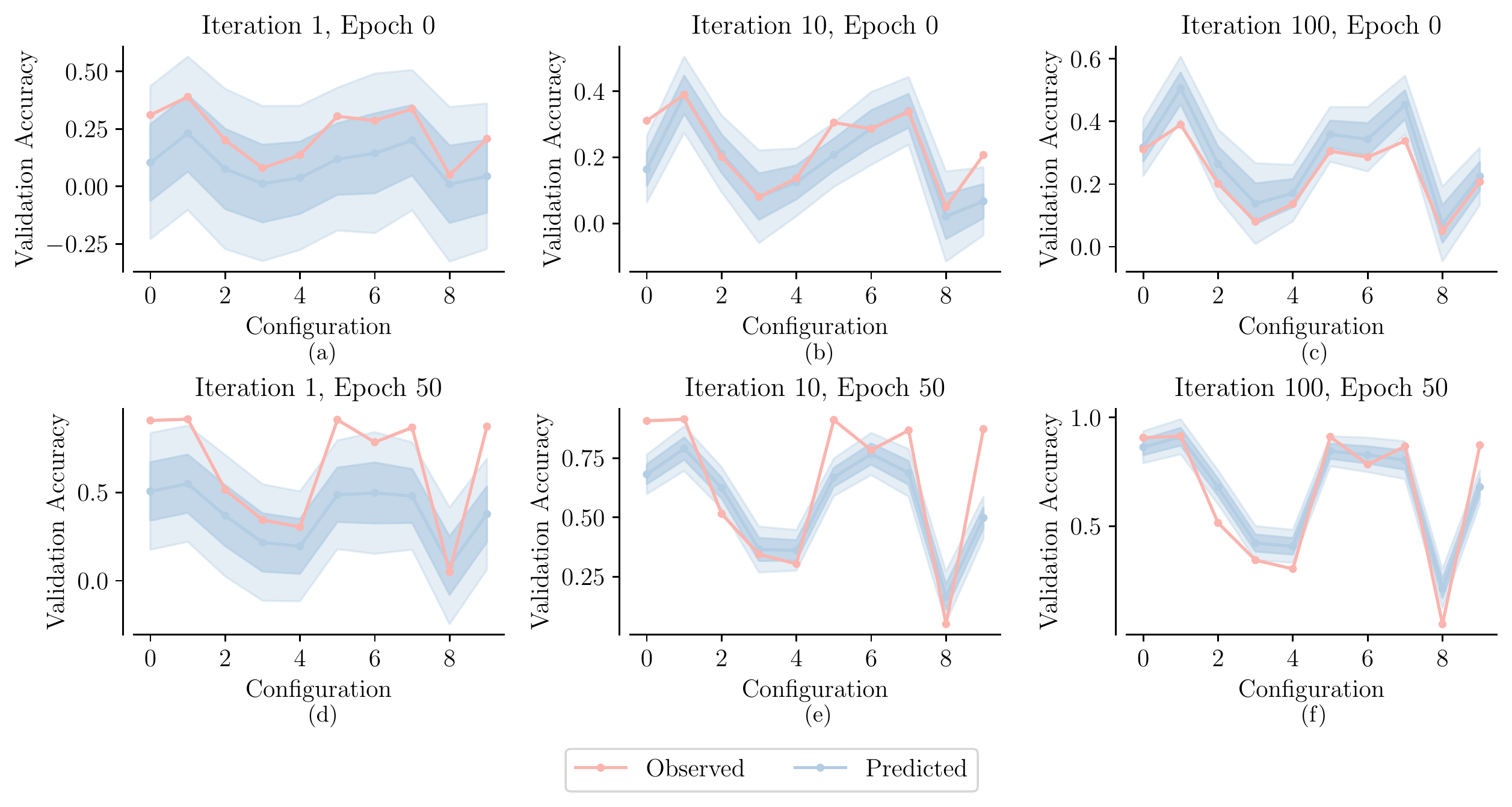}
    \caption{Predictions by \texttt{AT2} versus observed validation accuracies for one sampled train-test task pair and $10$ sampled configurations in \textrm{HyperRec}. The shaded regions indicate one and two predictive standard deviations. At iteration $1$, \texttt{AT2} approximates the optimization landscape of the new tuning task by transferring knowledge from past tasks. As more queries are made on the new task over iterations, \texttt{AT2} better resembles the observed validation accuracies at both low and high fidelities (i.e., Epoch $0$ and $50$, respectively).}
    \label{fig:qualitative}
\end{figure*}

\subsection{Amortized Auto-Tuning (\texttt{AT2}) Method} \label{sec:3.4}

Besides the kernel analysis, we propose Max-Trial-GP-UCB, a specially designed acquisition function. 
Similar to GP-UCB \cite{auer2002using, srinivas2010gaussian}, it defines an upper confidence bound $\phi(x) = \mu(x) + \eta \cdot \sigma(x)$ on $f(x)$ for a given $x$. 
Here, $\mu(x)$ and $\sigma(x)$ are the predictive mean and standard deviation, respectively, of the posterior distribution in Equation~\ref{eq:gp_posterior}. 
$\eta$ is a hyperparameter controlling the trade-off between exploration and exploitation. 
Since our ultimate goal is to identify the configuration with the highest final performance on the tuning task $t^*$, we choose the configuration to query at iteration $q$ as $c_q = \arg\max_{c\in\mathcal{C}} \max_{e\in\mathcal{E}} \phi((t^*, c, e))$.
Note that the fidelity space can only be queried in incremental order as described in Section~\ref{sec:2.1}.
Specifically, suppose that for $t^*$ and $c_q$, the maximum queried fidelity by the last iteration is $e_{\max}$, then we formulate the query at the current iteration as $x_q = (t^*, c_q, e_{\max}+1)$.
As a result, the computational cost involved in each query iteration is \textit{consistent} and equivalent to training a model with a hyperparameter configuration $c_q$ for a single epoch.
This contrasts with multi-fidelity UCB-based procedures which, during one iteration, train a model with a chosen configuration for multiple epochs up to the chosen (non-incremental) fidelity \cite{kandasamy2016gaussian}.

All together, we present the \underline{A}mor\underline{T}ized \underline{A}uto-\underline{T}uning (\texttt{AT2}) method as an instantiation of the multi-task multi-fidelity BO framework, consisting of \texttt{OptiLand} task kernel, \texttt{DeepPoly} configuration kernel, \texttt{AccCurve} fidelity kernel, and Max-Trial-GP-UCB acquisition function.
\texttt{AT2} measures inter-task similarity based on low-fidelity observations and enjoys the power of the best-performing kernel ensemble to transfer knowledge to new tuning tasks in a flexible and cost-efficient manner.

\section{Experiments} \label{sec:4}

\subsection{Experimental Setup} \label{sec:4.1}

\header{Datasets.} 
Besides \textrm{HyperRec}, we consider another similar database \textrm{LCBench} \cite{zimmer2020auto} in our experiments. 
We sample $100$ configurations from each database to construct the configuration space $\mathcal{C}$. Since training epochs are treated as fidelity values in our setting, we have $|\mathcal{E}|=75$ in \textrm{HyperRec} and $|\mathcal{E}|=52$ in \textrm{LCBench}. We further normalize the numerical hyperparameters based on their respective sampling distributions and take $y$ to be the \textit{top five validation accuracy} in \textrm{HyperRec} and the \textit{validation balanced accuracy} in \textrm{LCBench}.
To assess the generalizability of model performance, we randomly sample five train-test task pairs from each database where one pair consists of four train tasks and one test task.

\begin{figure*}[h!]
    \centering
    \includegraphics[width=0.79\textwidth]{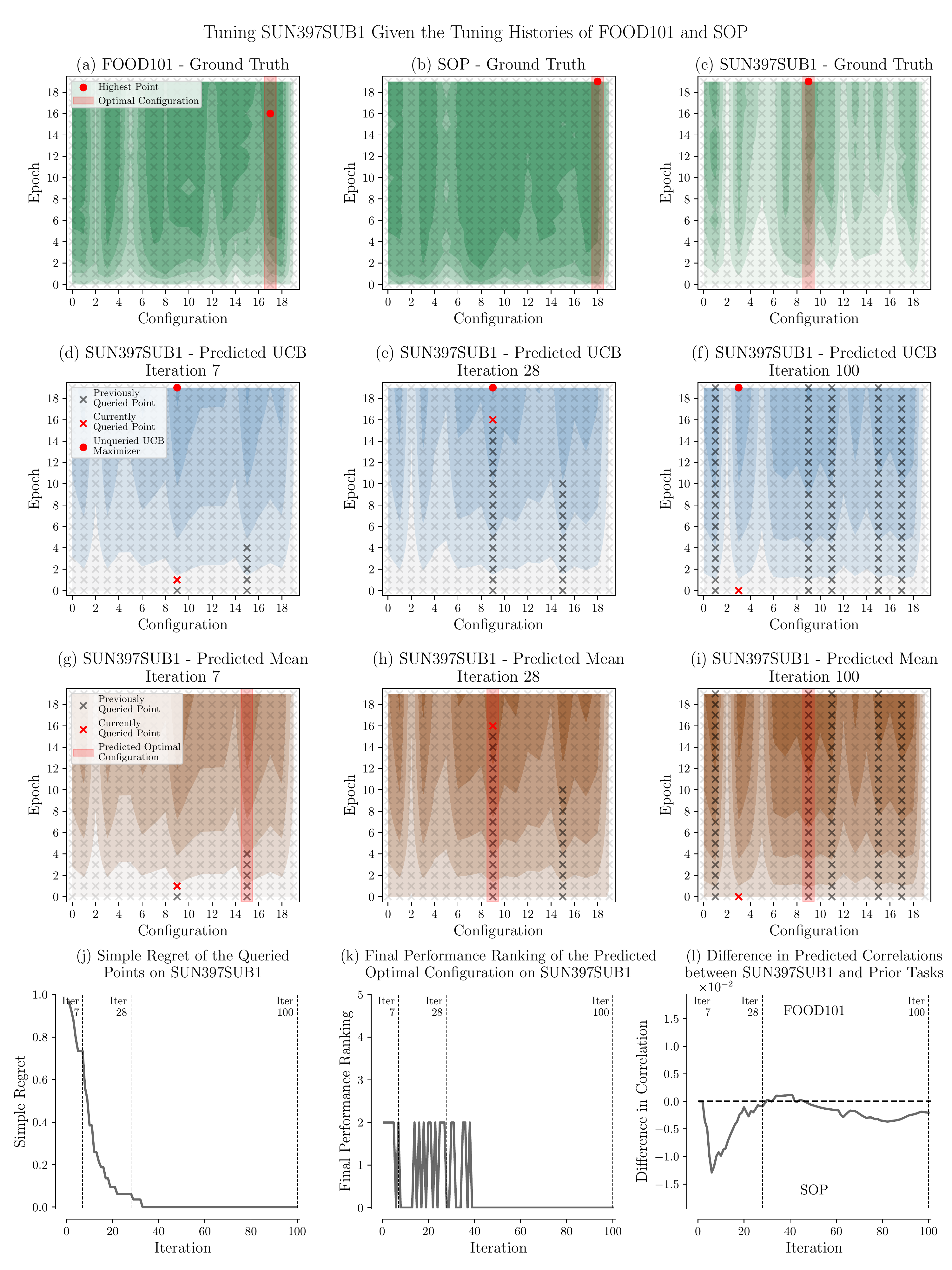}
    \caption{An example to demonstrate how \texttt{AT2} tunes a new task, \textrm{SUN397SUB1}, given the tuning histories of two prior tasks, \textrm{FOOD101} and \textrm{SOP}.
    For a clear presentation, we sample $20$ configurations and show $20$ epochs from \textrm{HyperRec} to form $\mathcal{C}$ and $\mathcal{E}$, respectively.
    Darker colors in the contour plots correspond to higher accuracies.
    At the beginning (i.e. Iteration 7), \texttt{AT2} makes a preliminary guess that Configuration 15 is the optimal configuration for \textrm{SUN397SUB1} via knowledge transfer.
    Later on (i.e. Iteration 28), \texttt{AT2} updates its belief based on cheap-to-obtain low-fidelity tuning observations and correctly recognizes the optimal configuration---Configuration 9.
    It continues to query Configuration 9 until the maximum fidelity and, consequently, reduces the simple regret to zero.}
    \label{fig:example}
\end{figure*}

\header{Baselines.} 
We compare our \texttt{AT2} method against a broad spectrum of hyperparameter transfer learning baselines: \texttt{ABLR} \cite{perrone2018scalable} applies Bayesian linear regression for each task with a shared representation space; \texttt{ABLR-HB} \cite{valkov2018simple} further utilizes \texttt{Hyperband} \cite{li2017hyperband} for multi-fidelity tuning; \texttt{Box-BO} \cite{perrone2019learning} constrains the search space of BO based on the best configurations of train tasks; \texttt{Box-BOHB} supplies \texttt{BOHB} \cite{falkner2018bohb} with a constrained search space to allow for inter-task knowledge transfer; \texttt{BOHAMIANN} \cite{springenberg2016bayesian} combines neural networks with stochastic gradient Hamiltonian Monte Carlo for better scalability; \texttt{PMF} \cite{fusi2018probabilistic} leverages probabilistic matrix factorization for hyperparameter recommendation; \texttt{RGPE} \cite{feurer2018scalable} ensembles single-task GPs as a ranking-weighted mixture. 
Moreover, \texttt{Box-BO} uses the same configuration and fidelity kernels as \texttt{AT2} for the BO process. \texttt{Box-BO} and \texttt{Box-BOHB} define the candidate pool based on the top three best-performing training configurations.

\header{Implementation details.}
We implement the proposed \texttt{AT2} method using the GPyTorch package \cite{gardner2018gpytorch}. More specifically, \texttt{AT2} is initialized with \numprint{1000} inducing points and optimized for $200$ epochs with the momentum optimizer (learning rate $=0.02$, momentum factor $=0.8$) and a linearly decaying scheduler. 
All the other baselines are trained with their respective default settings and incorporate fidelity value as a contextual variable so as to consider multi-fidelity information. 
We apply our novel Max-Trial-GP-UCB acquistion function to all the methods for selecting the next point to query and set $\eta=0.25$ to balance exploration and exploitation.
More details regarding the choice of hyperparameters are explained in Appendix~\ref{app:3}.

\subsection{Experimental Results} \label{sec:4.2}

\header{Quantitative evaluation.}
To assess the performance of baselines quantitatively, we allow a budget $Q$ of $100$ iterations and report the results averaged over five train-test task pairs for each database based on the two metrics ($R_Q$, $F_Q$) discussed in Section~\ref{sec:2.1}. 
Similar to Section~\ref{sec:2.3}, we use the final performance ranking for $F_Q$. 

As shown in Figure~\ref{fig:experiment}, although some methods give competitive simple regrets, their predicted final performance rankings deviate.
In particular, methods requiring preliminary observations for proper initialization (e.g., \texttt{BOHAMIANN} and \texttt{PMF}) generally do not work well under cold start situations.
The final performance rankings delivered by \texttt{ABLR} and \texttt{ABLR-HB} worsen after more observations become available, especially in \textrm{HyperRec}. This suggests that multi-fidelity information requires careful treatment as it is challenging to forecast max-fidelity performance. 
Moreover, while \texttt{Box-BO} shares the same configuration and fidelity kernels as \texttt{AT2}, it takes a much longer time for \texttt{Box-BO} to converge to a relatively satisfying result, which validates the significance of the task kernel in \texttt{AT2}.
In contrast, \texttt{RGPE} scores well for both metrics thanks to the performance of ensemble methods.

Last but not least, our proposed \texttt{AT2} method outperforms other baselines by a clear margin, improving the optimization quality (lower simple regrets) with lower computational cost (fewer iterations) and stabilizing on low final performance rankings even during the first few iterations.

\header{Qualitative analysis.} 
To better understand how effective \texttt{AT2} is in transferring knowledge and forecasting high-fidelity performance, we visualize how the predictive mean and standard deviation of the GP surrogate change over iterations.
We sample one train-test task pair and $10$ configurations from \textrm{HyperRec} and compare the predicted and observed accuracies in Figure~\ref{fig:qualitative}.

At the start when there are no observations on the new task, \texttt{AT2} exploits knowledge gained from prior tasks and produces a reasonable approximation of the true landscape albeit with high uncertainties.
As more queries are made on the new task over iterations, \texttt{AT2} can better extrapolate the high-fidelity performance (e.g., the zigzag shape from configuration 4 to 8) based on low-fidelity observations with a reduced standard deviation.

\header{Demonstrating example.}
To illustrate the multi-task multi-fidelity tuning process of \texttt{AT2}, we visualize an example of tuning a new task, \textrm{SUN397SUB1}, based on the tuning histories of two existing tasks, \textrm{FOOD101} and \textrm{SOP}, in Figure~\ref{fig:example}.
For a clear presentation, we sample $20$ configurations and show $20$ epochs from \textrm{HyperRec} to form $\mathcal{C}$ and $\mathcal{E}$, respectively.


In Figure~\ref{fig:example}, the first row displays the ground truth validation accuracies achieved by distinct configurations at various epochs in all three tasks.
The second row demonstrates how our Max-Trial-GP-UCB acquisition function determines the next point to query based on the predicted upper confidence bound (UCB) at three different time points during the tuning process (i.e., Iteration 7, Iteration 28, Iteration 100).
The third row shows the predicted optimal configuration based on the predicted mean at the same time points as the second row.
And the fourth row visualizes the change in three metrics over the tuning process, including simple regret, final performance ranking, and the difference in predicted correlations between \textrm{SUN397SUB1} and the existing tasks.

During the first few iterations, \texttt{AT2} predicts a higher correlation between \textrm{SUN397SUB1} and \textrm{SOP} based on the queried observations, as shown in Figure~\ref{fig:example} (l).
It leverages knowledge transfer and makes a preliminary guess in Figure~\ref{fig:example} (g) that Configuration 15 is the optimal configuration.
Since \texttt{AT2} exploits cheap-to-obtain low-fidelity tuning observations, it does not need to wait until querying the maximum fidelity of Configuration 15.
Instead, it forecasts in Figure~\ref{fig:example} (d) that (Configuration 9, Epoch 19) delivers the highest UCB and turns to query (Configuration 9, Epoch 1) at Iteration 7 due to the cost-efficiency design of the acquisition function (explained in Section~\ref{sec:3.4}).

As more observations become available afterward, \texttt{AT2} recalibrates the similarity between \textrm{SUN397SUB1} and \textrm{SOP} compared to that between \textrm{SUN397SUB1} and \textrm{FOOD101}, as shown in Figure~\ref{fig:example} (l).
It correctly identifies that Configuration 9 is the optimal one (Figure~\ref{fig:example} (h)) and continues to query Configuration 9 to the maximum fidelity (Figure~\ref{fig:example} (e)).
As a result, the simple regret drops to zero in Figure~\ref{fig:example} (j).
Later on, although more queries are made over iterations, \texttt{AT2} holds the correct belief that Configuration 9 is the optimal through the full iteration budget (Figures~\ref{fig:example} (i) and (k)).
\section{Conclusion} \label{sec:5}

In this paper, to achieve cost-efficient hyperparameter transfer optimization, we leverage cheap-to-obtain low-fidelity tuning observations for measuring inter-task dependencies.
Based on a systematic survey of $40$ existing baselines and a thorough analysis of a multi-task multi-fidelity BO framework, we propose the \underline{A}mor\underline{T}ized \underline{A}uto-\underline{T}uning (\texttt{AT2}) method. 
We further compute a \underline{Hyper}parameter \underline{Rec}ommendation (\textrm{HyperRec}) database offline to serve the community. 
The compelling empirical performance of our \texttt{AT2} method on \textrm{HyperRec} and other real-world databases demonstrates the method's effectiveness.
In the future, we plan to investigate how other surrogate models (e.g., random forest \cite{hutter2011sequential}) and acquisition functions (e.g., information-based \cite{takeno2020multi}) perform in our multi-task multi-fidelity BO framework.

\newpage
\bibliographystyle{IEEEtran}
\bibliography{main}

\newpage
\begin{IEEEbiography}
[{\includegraphics[width=1in,height=1.25in,clip,keepaspectratio]{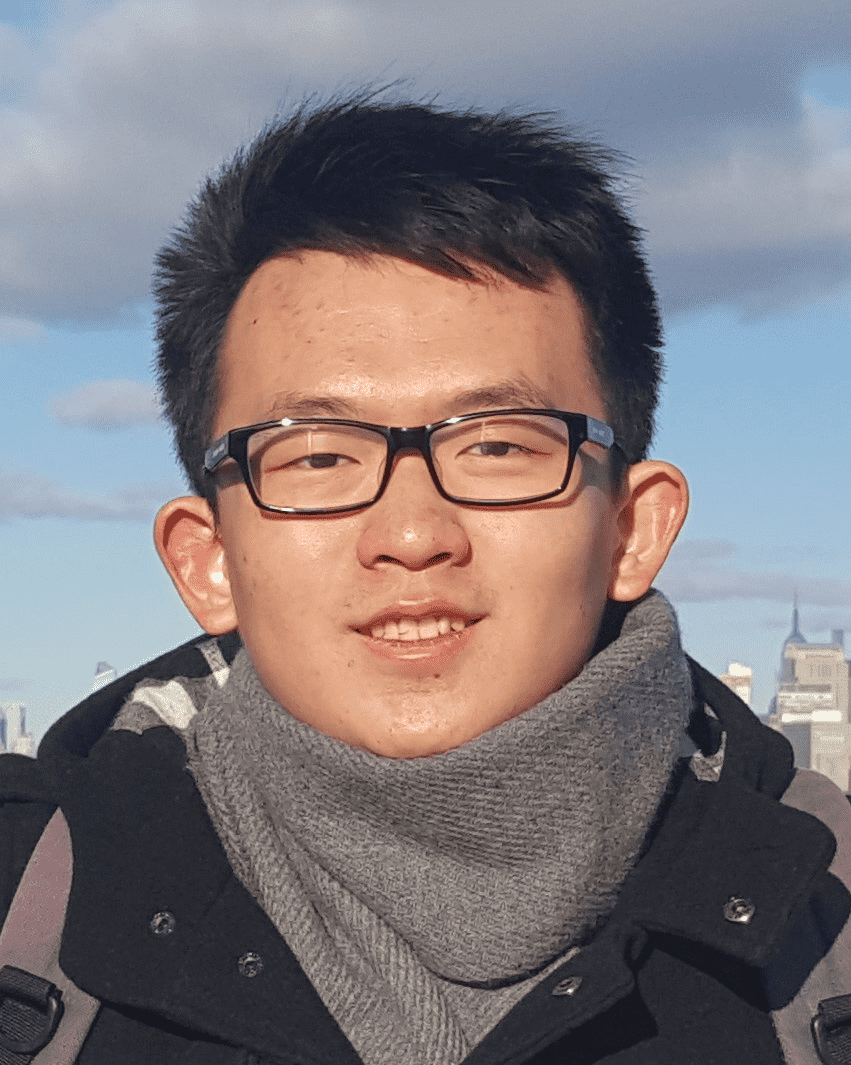}}]
{Yuxin Xiao} obtained his M.S. in Machine Learning at Carnegie Mellon University in 2022. He received his B.S. in Computer Science and B.S. in Statistics and Mathematics at the University of Illinois at Urbana-Champaign in 2020. He focuses on uncertainty-aware machine learning on structured data and has published first-authored papers in WWW, TKDE, and IEEE BigData. Yuxin also received the CRA Outstanding Undergraduate Researcher Award and C.W. Gear Outstanding Undergraduate Award at UIUC in 2020.
\end{IEEEbiography}

\vspace{-300pt}
\begin{IEEEbiography}
[{\includegraphics[width=1in,height=1.25in,clip,keepaspectratio]{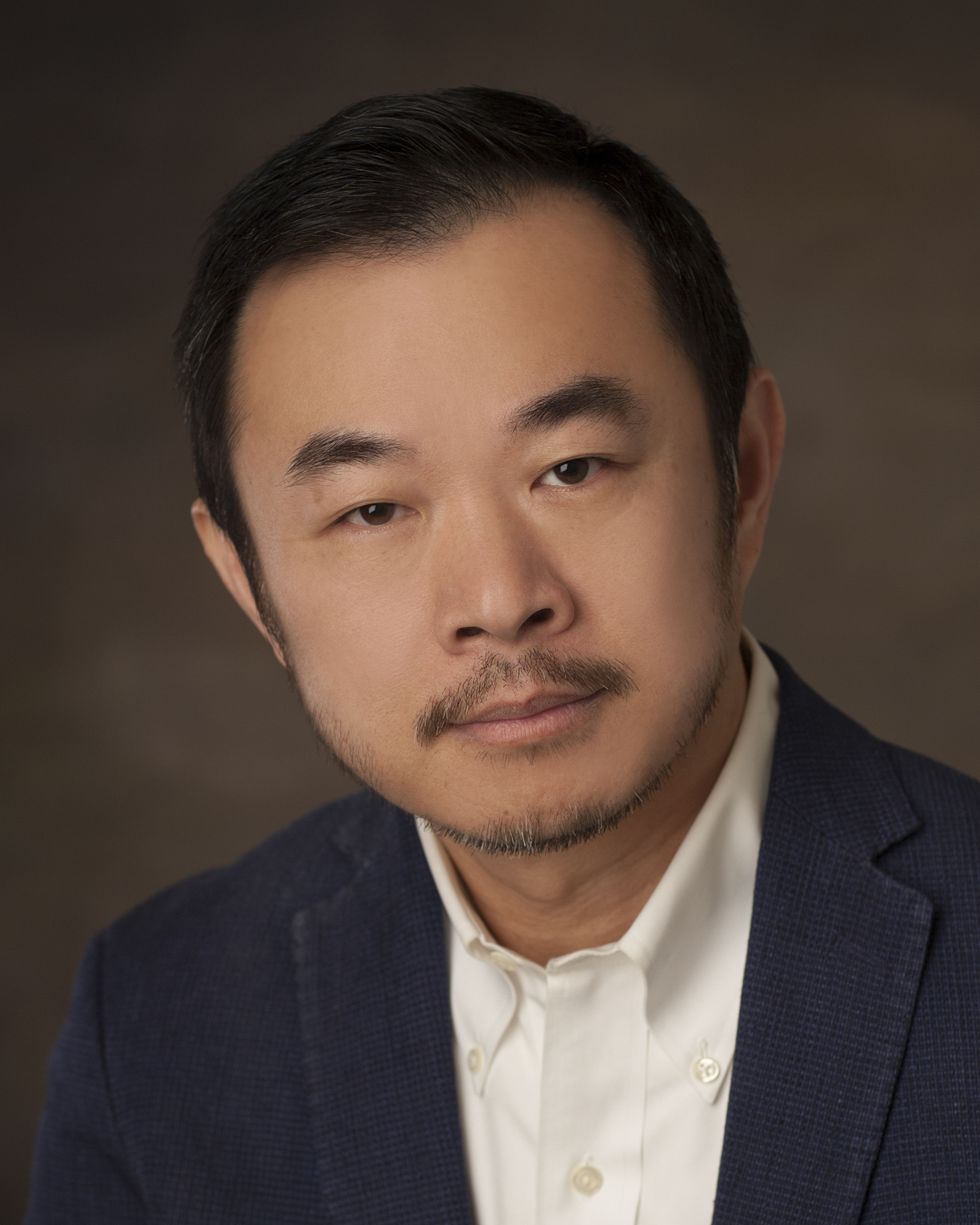}}]
{Eric P. Xing} is a Professor of Computer Science at Carnegie Mellon University, president of Mohamed bin Zayed University of Artificial Intelligence, and the Founder and Chief Scientist of Petuum, Inc.
He completed his undergraduate study at Tsinghua University, and holds a PhD in Molecular Biology and Biochemistry from the State University of New Jersey, and a PhD in Computer Science from the University of California, Berkeley. His main research interests are the development of machine learning and statistical methodology, and large-scale computational system and architectures, for solving problems involving automated learning, reasoning, and decision-making in high-dimensional, multimodal, and dynamic possible worlds in artificial, biological, and social systems.
He is a Fellow of the Association of Advancement of Artificial Intelligence (AAAI), and an IEEE Fellow.
\end{IEEEbiography}

\vspace{-300pt}
\begin{IEEEbiography}
[{\includegraphics[width=1in,height=1.25in,clip,keepaspectratio]{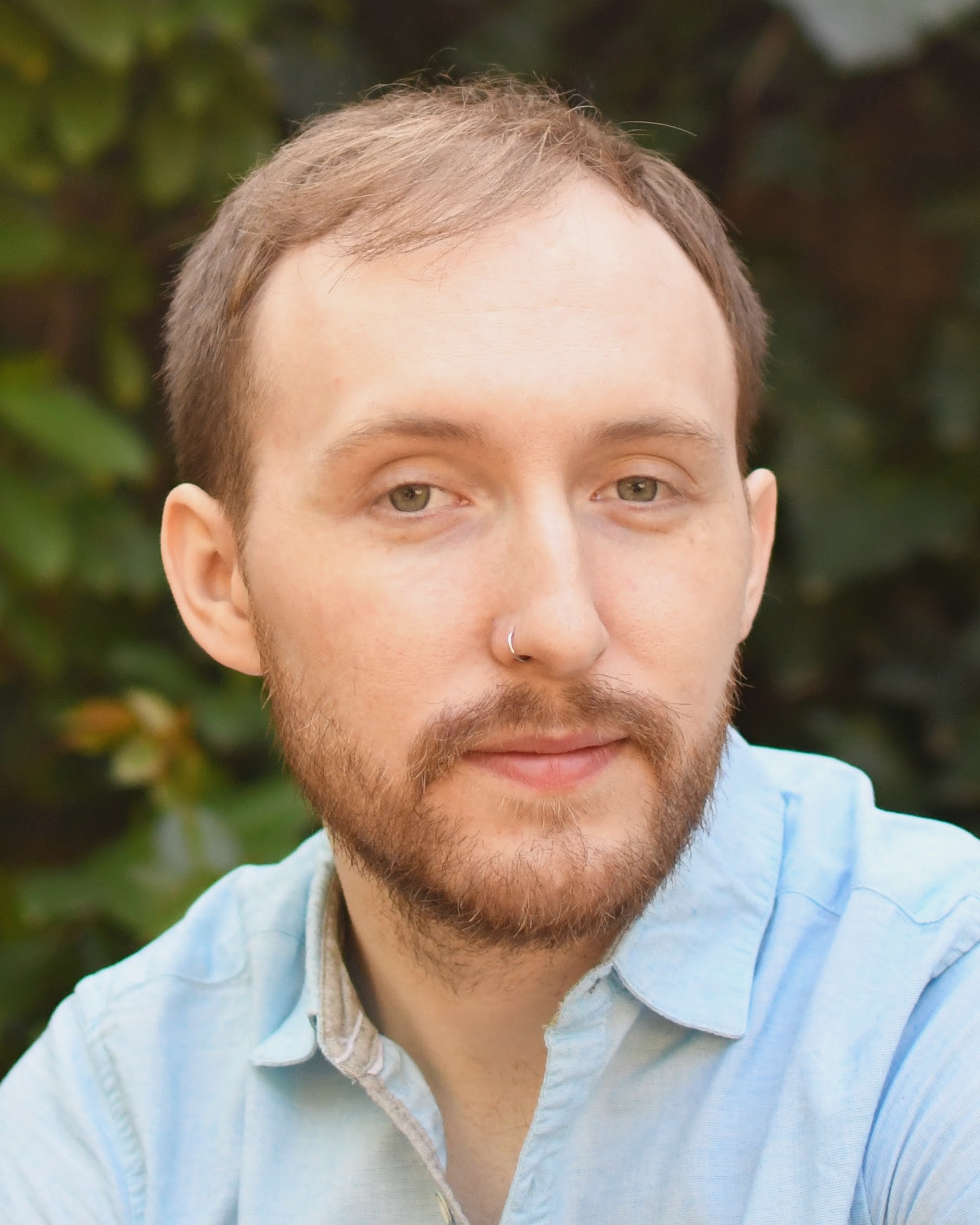}}]
{Willie Neiswanger} is a postdoctoral scholar at Stanford University. He completed his undergraduate study at Columbia University, and holds a Ph.D. in Machine Learning from Carnegie Mellon University.
His main research focus is on developing algorithms and systems to help scale and automate machine learning. He also works on uncertainty quantification, sequential decision making under uncertainty, and its application to problems in science and engineering.
\end{IEEEbiography}

\newpage
\onecolumn
\appendices
\section{Details of HyperRec Database} \label{app:1}

\subsection{Generation of \textrm{HyperRec}}\label{app:1.1}

The \underline{Hyper}parameter \underline{Rec}ommendation database (\textrm{HyperRec}) consists of 27 unique image classification tasks and 150 distinct configurations sampled from a 16-dimensional nested hyperparameter space. The original image classification dataset of each task is split based on a common ratio: $60\%$ for the training set, $20\%$ for the validation set, and $20\%$ for the testing set. We summarize the details of the tasks in Table~\ref{tab:tasks} and explain the nested hyperparameter space in Appendix~\ref{app:1.2}.

For each task, we evaluate each configuration during 75 training epochs and repeat this with two randomly sampled seeds. During training, we record the batch-wise cross-entropy loss, the batch-wise top one, five, and ten accuracies, and the training time taken to loop through all the batches. During evaluation, we record the epoch-wise cross-entropy loss and the epoch-wise top one, five, and ten accuracies for the validation and testing sets separately, as well as the evaluation time taken to loop through the two sets. 

\begin{table*}[h!]
  \centering
  \begin{tabular}{lll} 
    \toprule
    \textbf{Task}/\textbf{Dataset} & \textbf{Number of Images} & \textbf{Number of Classes} \\
    \midrule
    \midrule
\textrm{ACTION40} \cite{yao2011human}   & $\numprint{9532}$      & $\numprint{40}$    \\
\textrm{AWA2} \cite{xian2018zero}       & $\numprint{37322}$     & $\numprint{50}$    \\
\textrm{BOOKCOVER30} \cite{iwana2016judging}& $\numprint{57000}$     & $\numprint{30}$    \\
\textrm{CALTECH256} \cite{griffin2007caltech} & $\numprint{30607}$     & $\numprint{257}$    \\
\textrm{CARS196} \cite{KrauseStarkDengFei-Fei_3DRR2013}    & $\numprint{16185}$     & $\numprint{196}$    \\
\textrm{CIFAR10} \cite{krizhevsky2009learning}    & $\numprint{60000}$     & $\numprint{10}$    \\
\textrm{CIFAR100} \cite{krizhevsky2009learning}   & $\numprint{60000}$     & $\numprint{100}$   \\
\textrm{CUB200} \cite{WahCUB_200_2011}     & $\numprint{11788}$     & $\numprint{200}$    \\
\textrm{FLOWER102} \cite{nilsback2008automated}  & $\numprint{8189}$      & $\numprint{102}$   \\
\textrm{FOOD101} \cite{bossard14}    & $\numprint{101000}$    & $\numprint{101}$   \\
\textrm{IMAGENET64SUB1} \cite{ILSVRC15}   & $\numprint{128112}$     & $\numprint{1000}$    \\
\textrm{IMAGENET64SUB2} \cite{ILSVRC15}   & $\numprint{128112}$     & $\numprint{1000}$    \\
\textrm{IMAGENET64SUB3} \cite{ILSVRC15}   & $\numprint{128112}$     & $\numprint{1000}$    \\
\textrm{IP102} \cite{Wu2019Insect}      & $\numprint{75222}$     & $\numprint{102}$    \\
\textrm{ISR} \cite{quattoni2009recognizing}        & $\numprint{15620}$     & $\numprint{67}$    \\
\textrm{OIPETS} \cite{parkhi2012cats}     & $\numprint{7349}$      & $\numprint{37}$    \\
\textrm{PLACE365SUB1} \cite{zhou2017places}   & $\numprint{91987}$     & $\numprint{365}$    \\
\textrm{PLACE365SUB2} \cite{zhou2017places}   & $\numprint{91987}$     & $\numprint{365}$    \\
\textrm{PLACE365SUB3} \cite{zhou2017places}   & $\numprint{91987}$     & $\numprint{365}$    \\
\textrm{PLANT39} \cite{geetharamani2019identification}    & $\numprint{61486}$     & $\numprint{39}$    \\
\textrm{RESISC45} \cite{cheng2017remote}     & $\numprint{31500}$     & $\numprint{45}$    \\
\textrm{SCENE15} \cite{fei2005bayesian, lazebnik2006beyond}    & $\numprint{4485}$      & $\numprint{15}$    \\
\textrm{SDD} \cite{KhoslaYaoJayadevaprakashFeiFei_FGVC2011, imagenet_cvpr09}        & $\numprint{20580}$     & $\numprint{120}$    \\
\textrm{SOP} \cite{song2016deep}        & $\numprint{120053}$    & $\numprint{12}$    \\
\textrm{SUN397SUB1} \cite{xiao2010sun, xiao2016sun}   & $\numprint{9925}$     & $\numprint{397}$    \\
\textrm{SUN397SUB2} \cite{xiao2010sun, xiao2016sun}   & $\numprint{9925}$     & $\numprint{397}$    \\
\textrm{SUN397SUB3} \cite{xiao2010sun, xiao2016sun}   & $\numprint{9925}$     & $\numprint{397}$    \\
    \bottomrule 
  \end{tabular}
  \caption{Details on the 27 tasks in \textrm{HyperRec}.}
  \label{tab:tasks}
\end{table*}

\subsection{Nested Hyperparameter Space in \textrm{HyperRec}} \label{app:1.2}

Here, we explain the 16-dimensional nested hyperparameter space used in \textrm{HyperRec}. In what follows, $\mathcal{C}\{\cdots\}$ denotes the categorical distribution, $\mathcal{U}(\cdot,\cdot)$ denotes the uniform distribution, $\mathcal{U}\{\cdot,\cdot\}$ denotes the discrete uniform distribution, $\mathcal{LU}(\cdot,\cdot)$ denotes the log-uniform distribution, and \texttt{CAWR} stands for \texttt{CosineAnnealingWarmRestarts}.

In Table~\ref{tab:root_hyperparameters}, we summarize information about the subset of hyperparameters in \textrm{HyperRec} that are independent of any categorial variables.

\begin{table*}[h!]
  \centering
  \begin{tabular}{lcl} 
    \toprule
    \textbf{Hyperparameter} & & \textbf{Tuning Distribution} \\
    \midrule
    \midrule
    Batch size & & $\mathcal{U} \{32,\, 128\}$ \\
    Model & & $\mathcal{C} \{\texttt{ResNet34},\, \texttt{ResNet50}\}$ \\
    Optimizer & & $\mathcal{C} \{\texttt{Adam},\, \texttt{Momentum}\}$ \\
    LR Scheduler & & $\mathcal{C} \{\texttt{StepLR},\, \texttt{ExponentialLR},\, \texttt{CyclicLR},\, \texttt{CAWR} \}$ \\
    \bottomrule 
  \end{tabular}
  \caption{Details on the hyperparameters that are independent of any categorical variables in \textrm{HyperRec}.}
  \label{tab:root_hyperparameters}
\end{table*}

\textrm{HyperRec} involves three categorical hyperparameters: Model, Optimizer, and Learning Rate (LR) Scheduler. In particular, we consider two choices for Model (\texttt{ResNet34} and \texttt{ResNet50} \cite{he2016deep}), two choices for Optimizer (\texttt{Adam} \cite{kingma2014adam} and \texttt{Momentum} \cite{polyak1964some}), and four choices for LR Scheduler (\texttt{StepLR}, \texttt{ExponentialLR}, \texttt{CyclicLR} \cite{smith2017cyclical}, \texttt{CAWR} \cite{loshchilov2016sgdr}). 
The dependent hyperparameters of the categorical variables Optimizer and LR Scheduler in \textrm{HyperRec} are described in Table~\ref{tab:optimizer_hyperparameters} and Table~\ref{tab:scheduler_hyperparameters}, respectively. Note that the categorical variable Model does not have any dependent hyperparameters in \textrm{HyperRec}.

\begin{table*}[h!]
  \centering
  \begin{tabular}{lclcl} 
\toprule
\textbf{Optimizer Choice} & & \textbf{Hyperparameter} & & \textbf{Tuning Distribution} \\
\midrule
\midrule
\multirow{4}{*}{\texttt{Adam}} & & Learning rate & & $\mathcal{LU} (10^{-4},\, 10^{-1})$ \\
 & & Weight decay & & $\mathcal{LU} (10^{-5},\, 10^{-3})$ \\
 & & $\text{Beta}_0$ & & $\mathcal{LU} (0.5,\, 0.999)$ \\
 & & $\text{Beta}_1$ & & $\mathcal{LU} (0.8,\, 0.999)$ \\
\midrule
\multirow{3}{*}{\texttt{Momentum}} & & Learning rate & & $\mathcal{LU} (10^{-4},\, 10^{-1})$ \\
 & & Weight decay & & $\mathcal{LU} (10^{-5},\, 10^{-3})$ \\
 & &  Momentum factor & & $\mathcal{LU} (10^{-3},\, 1)$ \\
\bottomrule
  \end{tabular}
  \caption{Details on the hyperparameters that are dependent on Optimizer choices in \textrm{HyperRec}.}
  \label{tab:optimizer_hyperparameters}
\end{table*}

\begin{table*}[h!]
  \centering
  \begin{tabular}{lclcl} 
\toprule
\textbf{LR Scheduler Choice} & & \textbf{Hyperparameter} & & \textbf{Tuning Distribution} \\
\midrule
\midrule
\multirow{2}{*}{\texttt{StepLR}} & & Step size & & $\mathcal{U} \{2,\, 20\}$ \\
 & & Gamma & & $\mathcal{LU} (0.1,\, 0.5)$ \\
\midrule
\texttt{ExponentialLR} & & Gamma & & $\mathcal{LU} (0.85,\, 0.999)$ \\
\midrule
\multirow{3}{*}{\texttt{CyclicLR}} & & Gamma & & $\mathcal{LU} (0.1,\, 0.5)$ \\
 & & Max learning rate & & $\min(1,\, \text{LR} \times \mathcal{U} (1.1,\, 1.5))$ \\
 & & Step size up & & $\mathcal{U} \{1,\, 10\}$ \\
\midrule
\multirow{3}{*}{\texttt{CAWR}} & & $\text{T}_0$ & & $\mathcal{U} \{2,\, 20\}$ \\
 & & $\text{T}_{\text{mult}}$ & & $\mathcal{U} \{1,\, 4\}$ \\
 & & $\text{Eta}_{\min}$ & & $\text{LR} \times \mathcal{U} (0.5,\, 0.9)$ \\
\bottomrule
  \end{tabular}
  \caption{Details on the hyperparameters that are dependent on LR Scheduler choices in \textrm{HyperRec}.}
  \label{tab:scheduler_hyperparameters}
\end{table*}

\subsection{Comparison of \textrm{HyperRec} against \textrm{LCBench}} \label{app:1.3}

To highlight the uniqueness of \textrm{HyperRec}, we compare it against the other database used in our experiments (i.e., \textrm{LCBench} \cite{zimmer2020auto}) in terms of evaluation tasks and hyperparameter space.

\begin{description}
    \item [\textbf{Evaluation Tasks}:] \textrm{HyperRec} is intentionally designed to focus on image classification tasks so as to better contribute to the computer vision community. Therefore, it features popular large-scale image classification tasks such as \textrm{IMAGENET64} \cite{ILSVRC15}, \textrm{PLACE365} \cite{zhou2017places}, \textrm{FOOD101} \cite{bossard14} , \textrm{SUN397} \cite{xiao2010sun, xiao2016sun}, etc. These tasks typically contain hundreds of classes (e.g., 1,000 classes in \textrm{IMAGENET64}) and high-resolution images (e.g., 120,000 pixels per image in \textrm{SUN397}). These characteristics impose unique challenges and require careful treatment when tuning hyperparameters for modern image classification tasks. On the other hand, \textrm{LCBench} evaluates tabular datasets from the AutoML Benchmark \cite{gijsbers2019open}. Those datasets usually contain much fewer classes (16 classes per dataset on average) and features (166 dimensions per instance on average).
    \item [\textbf{Hyperparameter Space}:] Modern machine learning pipelines generally involve nested hyperparameter spaces. For instance, the momentum factor is a hyperparameter unique to the momentum optimizer. Hence, we design a 16-dimensional nested hyperparameter space that includes both numerical (e.g., batch size) and categorical (e.g., optimizer) hyperparameters. Furthermore, to make the tuning results more practically useful to the computer vision community, we assess each sampled hyperparameter configuration based on the widely adopted \texttt{ResNet} \cite{he2016deep} family. In contrast, \textrm{LCBench} uses a 7-dimensional flat hyperparameter space that only considers numerical hyperparameters and fully connected neural networks.
\end{description}

By generating \textrm{HyperRec}, we believe that the hyperparameter optimization community can leverage it to test the effectiveness of different tuning methods by checking how fast a tuning method can identify good-performing hyperparameter configurations. Moreover, the computer vision community can fairly compare the performance of image classification models against those used in \textrm{HyperRec} (e.g., \texttt{ResNet50}).

\clearpage
\section{Analysis Results of Kernel Combinations} \label{app:2}

Based on the four options proposed for each of the three component kernels (task, configuration, and fidelity kernels) in Section~\ref{sec:3.3}, we empirically assess the resulting $64$ combinations via both quantitative and qualitative measures. We discuss the evaluation results in Section~\ref{sec:3.3.4} and present shorter versions of the following tables and figures in Table~\ref{tab:analysis} and Figure~\ref{fig:analysis}, respectively.

\begin{table}[h!]
  \centering
  \begin{tabular}{c|c|c|c|c|c} 
    \toprule
        \textbf{Rank} & \textbf{Figure} & \textbf{Task Kernel} & \textbf{Configuration Kernel} & \textbf{Fidelity Kernel} & \textbf{ELBO} \\
    \midrule
    \midrule
        1 & Figure~\ref{fig:analysis_0} (a) & \texttt{OptiLand} & \texttt{DeepPoly} & \texttt{AccCurve} & $1.4951$ \\
        2 & Figure~\ref{fig:analysis_0} (b) & \texttt{MTBO} & \texttt{DeepLinear} & \texttt{Matern} & $1.4296$ \\
        3 & Figure~\ref{fig:analysis_0} (c) & \texttt{OptiLand} & \texttt{DeepLinear} & \texttt{AccCurve} & $1.4172$ \\
        4 & Figure~\ref{fig:analysis_0} (d) & \texttt{OptiLand} & \texttt{DeepPoly} & \texttt{Matern} & $1.4127$ \\
        5 & Figure~\ref{fig:analysis_0} (e) & \texttt{MTBO} & \texttt{DeepPoly} & \texttt{AccCurve} & $1.4121$ \\
        6 & Figure~\ref{fig:analysis_0} (f) & \texttt{DeepPoly} & \texttt{DeepLinear} & \texttt{AccCurve} & $1.4017$ \\
        7 & Figure~\ref{fig:analysis_0} (g) & \texttt{OptiLand} & \texttt{DeepLinear} & \texttt{Matern} & $1.3871$ \\
        8 & Figure~\ref{fig:analysis_0} (h) & \texttt{DeepLinear} & \texttt{DeepPoly} & \texttt{RBF} & $1.3763$ \\
        9 & Figure~\ref{fig:analysis_0} (i) & \texttt{DeepPoly} & \texttt{DeepLinear} & \texttt{Matern} & $1.3700$ \\
    \bottomrule
  \end{tabular}
  \caption{Quantitative performance of different kernel compositions (ranking $\#1\sim\#9$).}
  \label{tab:analysis_0}
\end{table}

\begin{figure}[h!]
    \centering
    \includegraphics[width=0.87\textwidth]{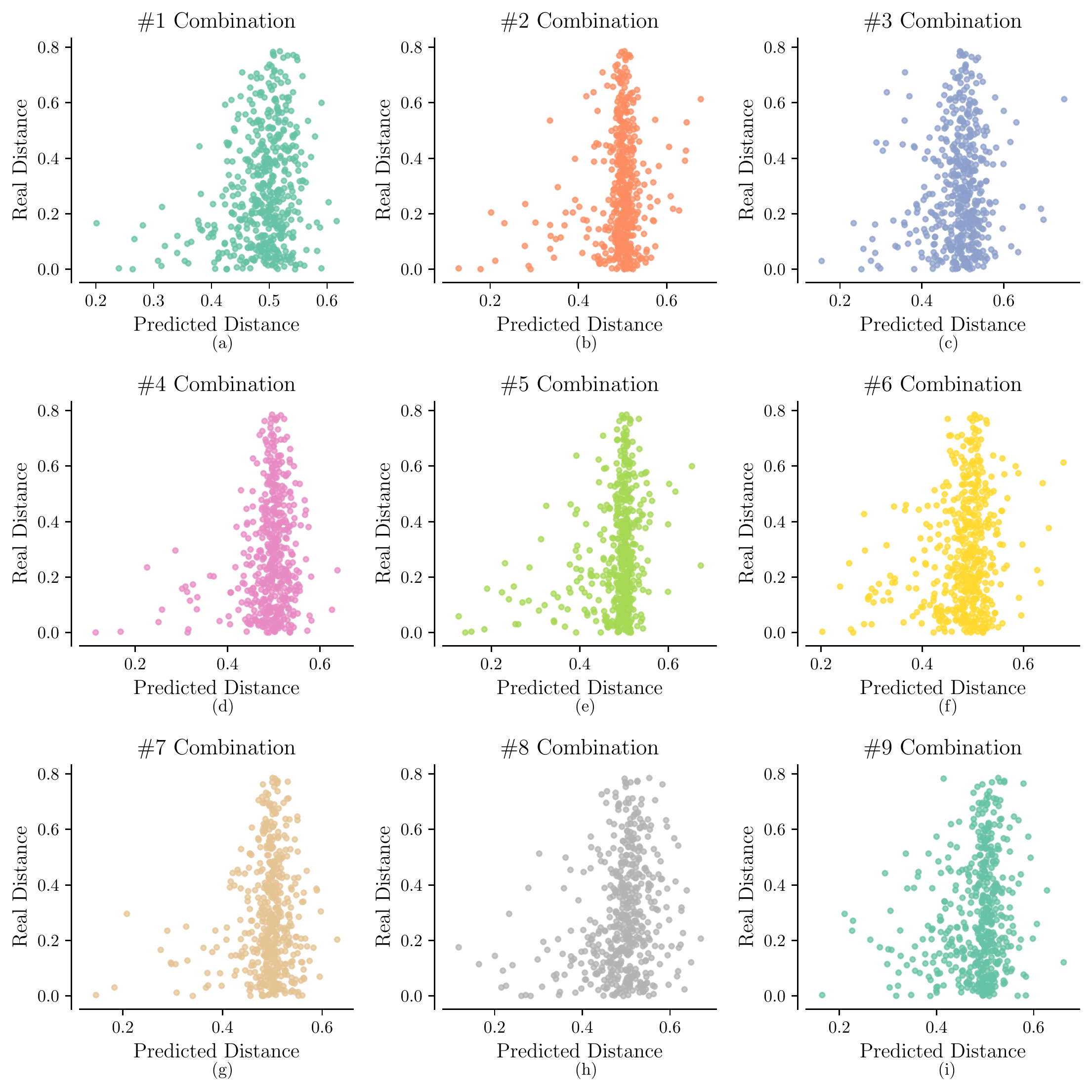}
    \caption{Qualitative performance of different kernel compositions (ranking $\#1\sim\#9$).}
    \label{fig:analysis_0}
\end{figure}

\clearpage

\begin{table*}[h!]
  \centering
  \begin{tabular}{c|c|c|c|c|c} 
    \toprule
        \textbf{Rank} & \textbf{Figure} & \textbf{Task Kernel} & \textbf{Configuration Kernel} & \textbf{Fidelity Kernel} & \textbf{ELBO} \\
    \midrule
    \midrule
        10 & Figure~\ref{fig:analysis_1} (a) & \texttt{OptiLand} & \texttt{DeepLinear} & \texttt{RBF} & $1.3590$ \\
        11 & Figure~\ref{fig:analysis_1} (b) & \texttt{DeepLinear} & \texttt{DeepLinear} & \texttt{RBF} & $1.3586$ \\
        12 & Figure~\ref{fig:analysis_1} (c) & \texttt{DeepLinear} & \texttt{DeepLinear} & \texttt{Matern} & $1.3464$ \\
        13 & Figure~\ref{fig:analysis_1} (d) & \texttt{DeepPoly} & \texttt{DeepLinear} & \texttt{RBF} & $1.3460$ \\
        14 & Figure~\ref{fig:analysis_1} (e) & \texttt{DeepPoly} & \texttt{DeepPoly} & \texttt{RBF} & $1.3360$ \\
        15 & Figure~\ref{fig:analysis_1} (f) & \texttt{MTBO} & \texttt{DeepLinear} & \texttt{AccCurve} & $1.3350$ \\
        16 & Figure~\ref{fig:analysis_1} (g) & \texttt{DeepLinear} & \texttt{DeepLinear} & \texttt{AccCurve} & $1.3309$ \\
        17 & Figure~\ref{fig:analysis_1} (h) & \texttt{MTBO} & \texttt{DeepLinear} & \texttt{RBF} & $1.3170$ \\
        18 & Figure~\ref{fig:analysis_1} (i) & \texttt{OptiLand} & \texttt{Tree} & \texttt{AccCurve} & $1.3129$ \\
    \bottomrule
  \end{tabular}
  \caption{Quantitative performance of different kernel compositions (ranking $\#10\sim\#18$).}
  \label{tab:analysis_1}
\end{table*}

\begin{figure*}[h!]
    \centering
    \includegraphics[width=0.87\textwidth]{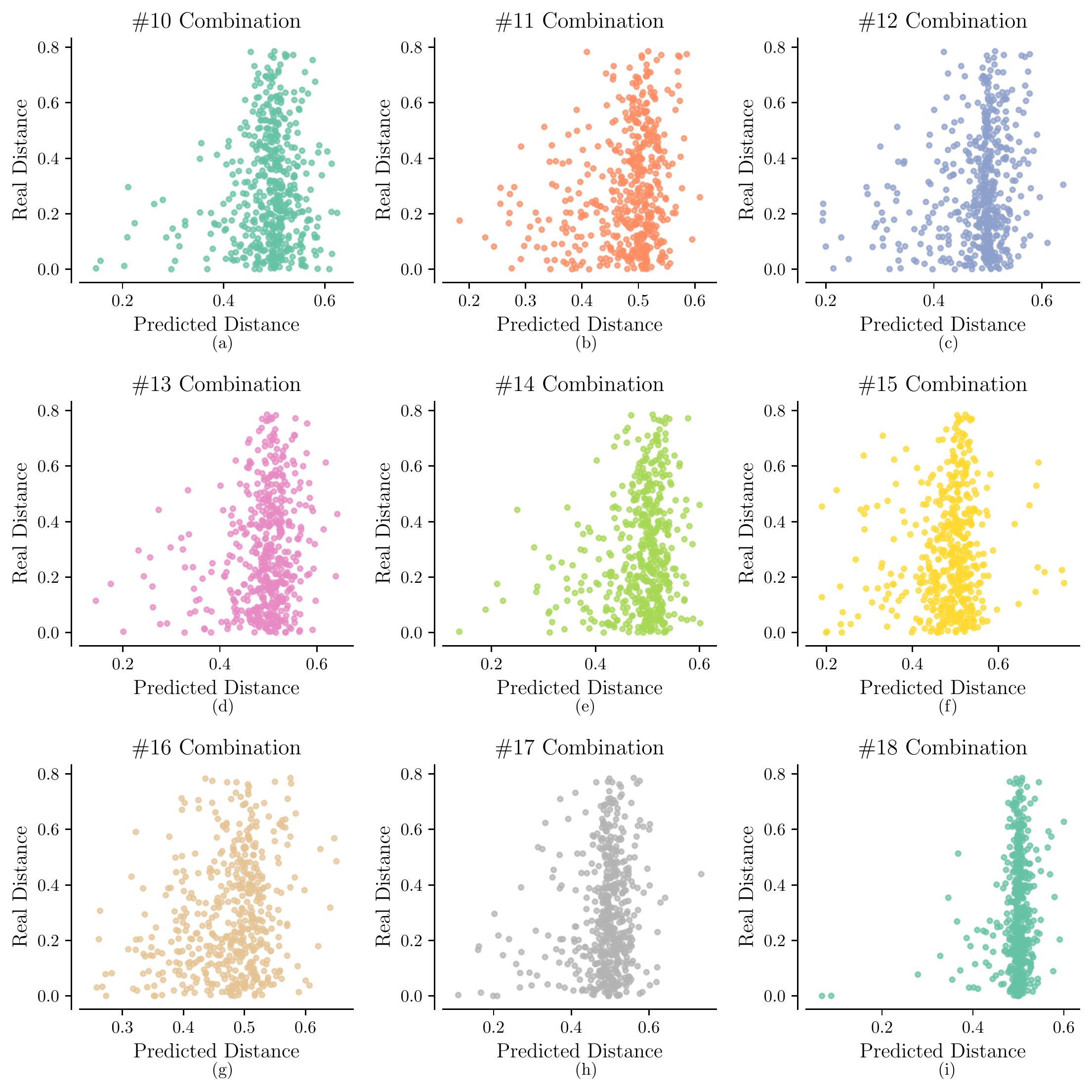}
    \caption{Qualitative performance of different kernel compositions (ranking $\#10\sim\#18$).}
    \label{fig:analysis_1}
\end{figure*}

\clearpage

\begin{table*}[h!]
  \centering
  \begin{tabular}{c|c|c|c|c|c} 
    \toprule
        \textbf{Rank} & \textbf{Figure} & \textbf{Task Kernel} & \textbf{Configuration Kernel} & \textbf{Fidelity Kernel} & \textbf{ELBO} \\
    \midrule
    \midrule
        19 & Figure~\ref{fig:analysis_2} (a) & \texttt{OptiLand} & \texttt{DeepPoly} & \texttt{RBF} & $1.2934$ \\
        20 & Figure~\ref{fig:analysis_2} (b) & \texttt{DeepPoly} & \texttt{DeepPoly} & \texttt{AccCurve} & $1.2925$ \\
        21 & Figure~\ref{fig:analysis_2} (c) & \texttt{DeepLinear} & \texttt{DeepPoly} & \texttt{AccCurve} & $1.2861$ \\
        22 & Figure~\ref{fig:analysis_2} (d) & \texttt{DeepLinear} & \texttt{Tree} & \texttt{AccCurve} & $1.2777$ \\
        23 & Figure~\ref{fig:analysis_2} (e) & \texttt{MTBO} & \texttt{DeepPoly} & \texttt{Matern} & $1.2763$ \\
        24 & Figure~\ref{fig:analysis_2} (f) & \texttt{DeepLinear} & \texttt{DeepPoly} & \texttt{Matern} & $1.2716$ \\
        25 & Figure~\ref{fig:analysis_2} (g) & \texttt{DeepPoly} & \texttt{Tree} & \texttt{AccCurve} & $1.2704$ \\
        26 & Figure~\ref{fig:analysis_2} (h) & \texttt{MTBO} & \texttt{DeepPoly} & \texttt{RBF} & $1.2495$ \\
        27 & Figure~\ref{fig:analysis_2} (i) & \texttt{OptiLand} & \texttt{Flat} & \texttt{AccCurve} & $1.2409$ \\
    \bottomrule
  \end{tabular}
  \caption{Quantitative performance of different kernel compositions (ranking $\#19\sim\#27$).}
  \label{tab:analysis_2}
\end{table*}

\begin{figure*}[h!]
    \centering
    \includegraphics[width=0.87\textwidth]{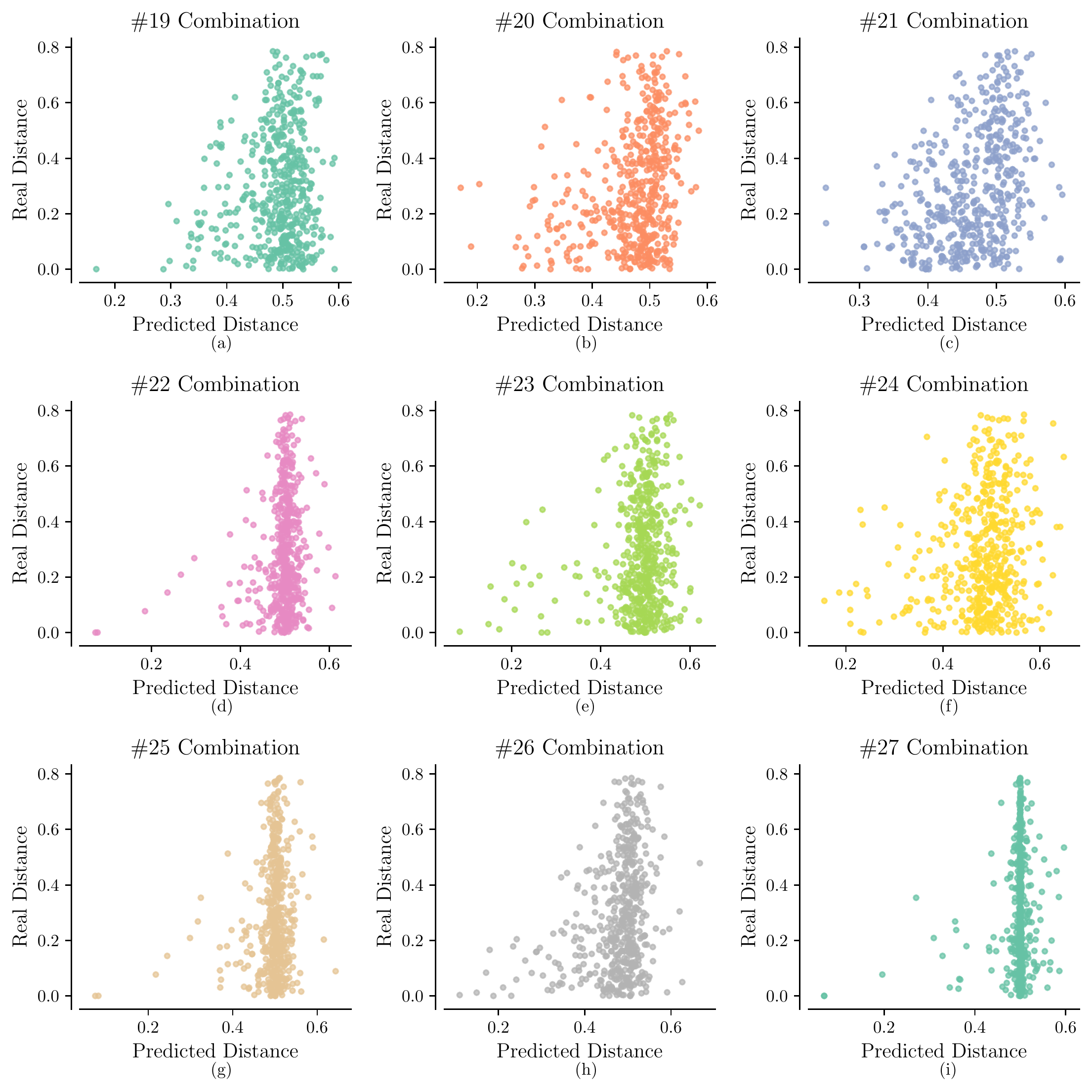}
    \caption{Qualitative performance of different kernel compositions (ranking $\#19\sim\#27$).}
    \label{fig:analysis_2}
\end{figure*}

\clearpage

\begin{table*}[h!]
  \centering
  \begin{tabular}{c|c|c|c|c|c} 
    \toprule
        \textbf{Rank} & \textbf{Figure} & \textbf{Task Kernel} & \textbf{Configuration Kernel} & \textbf{Fidelity Kernel} & \textbf{ELBO} \\
    \midrule
    \midrule
        28 & Figure~\ref{fig:analysis_3} (a) & \texttt{MTBO} & \texttt{Tree} & \texttt{AccCurve} & $1.2121$ \\
        29 & Figure~\ref{fig:analysis_3} (b) & \texttt{OptiLand} & \texttt{Tree} & \texttt{Matern} & $1.1843$ \\
        30 & Figure~\ref{fig:analysis_3} (c) & \texttt{OptiLand} & \texttt{Tree} & \texttt{RBF} & $1.1628$ \\
        31 & Figure~\ref{fig:analysis_3} (d) & \texttt{DeepPoly} & \texttt{Tree} & \texttt{Matern} & $1.1532$ \\
        32 & Figure~\ref{fig:analysis_3} (e) & \texttt{DeepPoly} & \texttt{Tree} & \texttt{RBF} & $1.1431$ \\
        33 & Figure~\ref{fig:analysis_3} (f) & \texttt{MTBO} & \texttt{Tree} & \texttt{RBF} & $1.0923$ \\
        34 & Figure~\ref{fig:analysis_3} (g) & \texttt{MTBO} & \texttt{Tree} & \texttt{Matern} & $1.0864$ \\
        35 & Figure~\ref{fig:analysis_3} (h) & \texttt{DeepLinear} & \texttt{Tree} & \texttt{Matern} & $1.0861$ \\
        36 & Figure~\ref{fig:analysis_3} (i) & \texttt{DeepLinear} & \texttt{Tree} & \texttt{RBF} & $1.0850$ \\
    \bottomrule
  \end{tabular}
  \caption{Quantitative performance of different kernel compositions (ranking $\#28\sim\#36$).}
  \label{tab:analysis_3}
\end{table*}

\begin{figure*}[h!]
    \centering
    \includegraphics[width=0.87\textwidth]{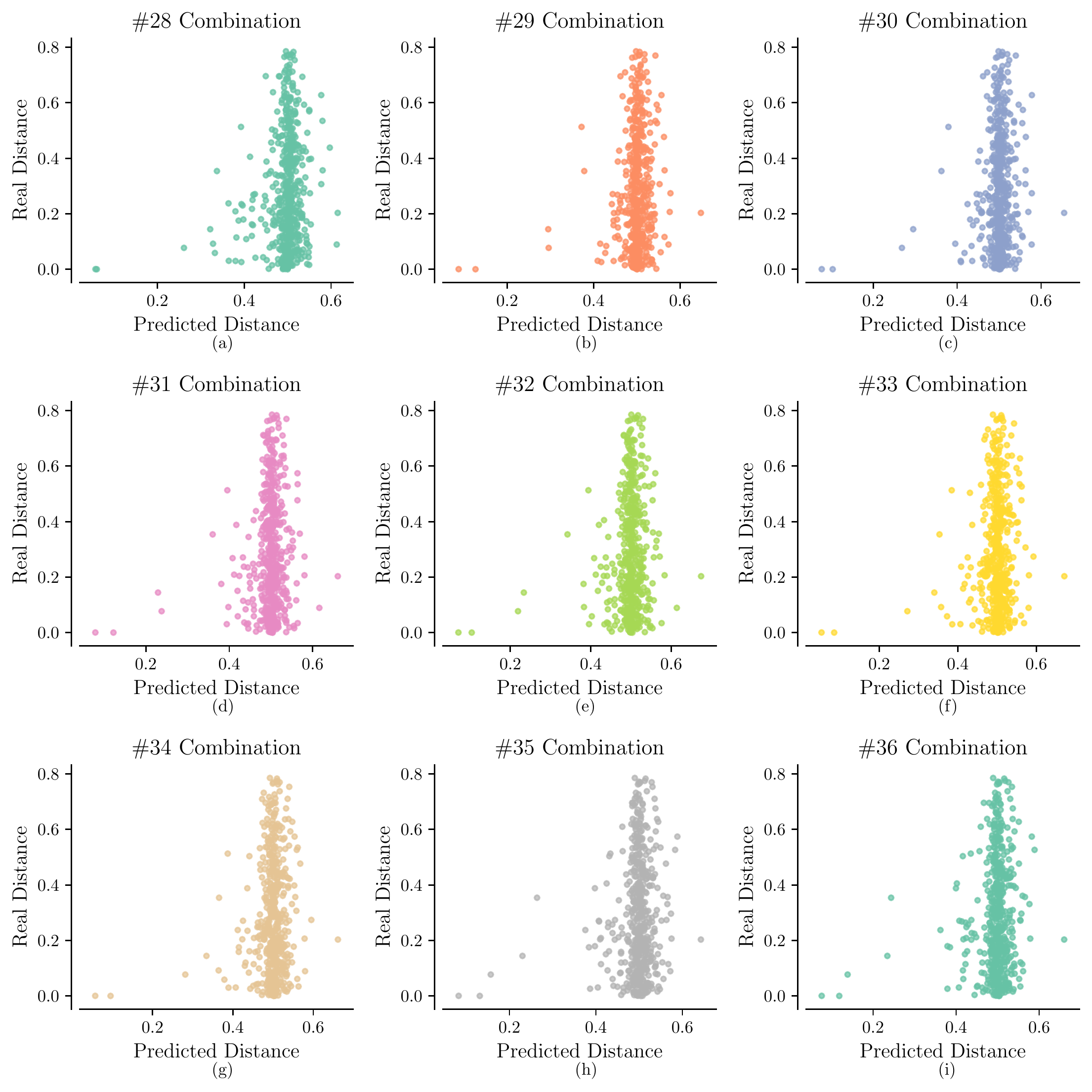}
    \caption{Qualitative performance of different kernel compositions (ranking $\#28\sim\#36$).}
    \label{fig:analysis_3}
\end{figure*}

\clearpage

\begin{table*}[h!]
  \centering
  \begin{tabular}{c|c|c|c|c|c} 
    \toprule
        \textbf{Rank} & \textbf{Figure} & \textbf{Task Kernel} & \textbf{Configuration Kernel} & \textbf{Fidelity Kernel} & \textbf{ELBO} \\
    \midrule
    \midrule
        37 & Figure~\ref{fig:analysis_4} (a) & \texttt{OptiLand} & \texttt{Flat} & \texttt{RBF} & $1.0633$ \\
        38 & Figure~\ref{fig:analysis_4} (b) & \texttt{OptiLand} & \texttt{Flat} & \texttt{Matern} & $1.0557$ \\
        39 & Figure~\ref{fig:analysis_4} (c) & \texttt{OptiLand} & \texttt{DeepPoly} & \texttt{Fabolas} & $1.0424$ \\
        40 & Figure~\ref{fig:analysis_4} (d) & \texttt{DeepPoly} & \texttt{Flat} & \texttt{AccCurve} & $1.0389$ \\
        41 & Figure~\ref{fig:analysis_4} (e) & \texttt{DeepPoly} & \texttt{DeepLinear} & \texttt{Fabolas} & $0.9881$ \\
        42 & Figure~\ref{fig:analysis_4} (f) & \texttt{DeepLinear} & \texttt{DeepPoly} & \texttt{Fabolas} & $0.9880$ \\
        43 & Figure~\ref{fig:analysis_4} (g) & \texttt{OptiLand} & \texttt{Tree} & \texttt{Fabolas} & $0.9224$ \\
        44 & Figure~\ref{fig:analysis_4} (h) & \texttt{OptiLand} & \texttt{Flat} & \texttt{Fabolas} & $0.9170$ \\
        45 & Figure~\ref{fig:analysis_4} (i) & \texttt{DeepPoly} & \texttt{Flat} & \texttt{RBF} & $0.9100$ \\
    \bottomrule
  \end{tabular}
  \caption{Quantitative performance of different kernel compositions (ranking $\#37\sim\#45$).}
  \label{tab:analysis_4}
\end{table*}

\begin{figure*}[h!]
    \centering
    \includegraphics[width=0.87\textwidth]{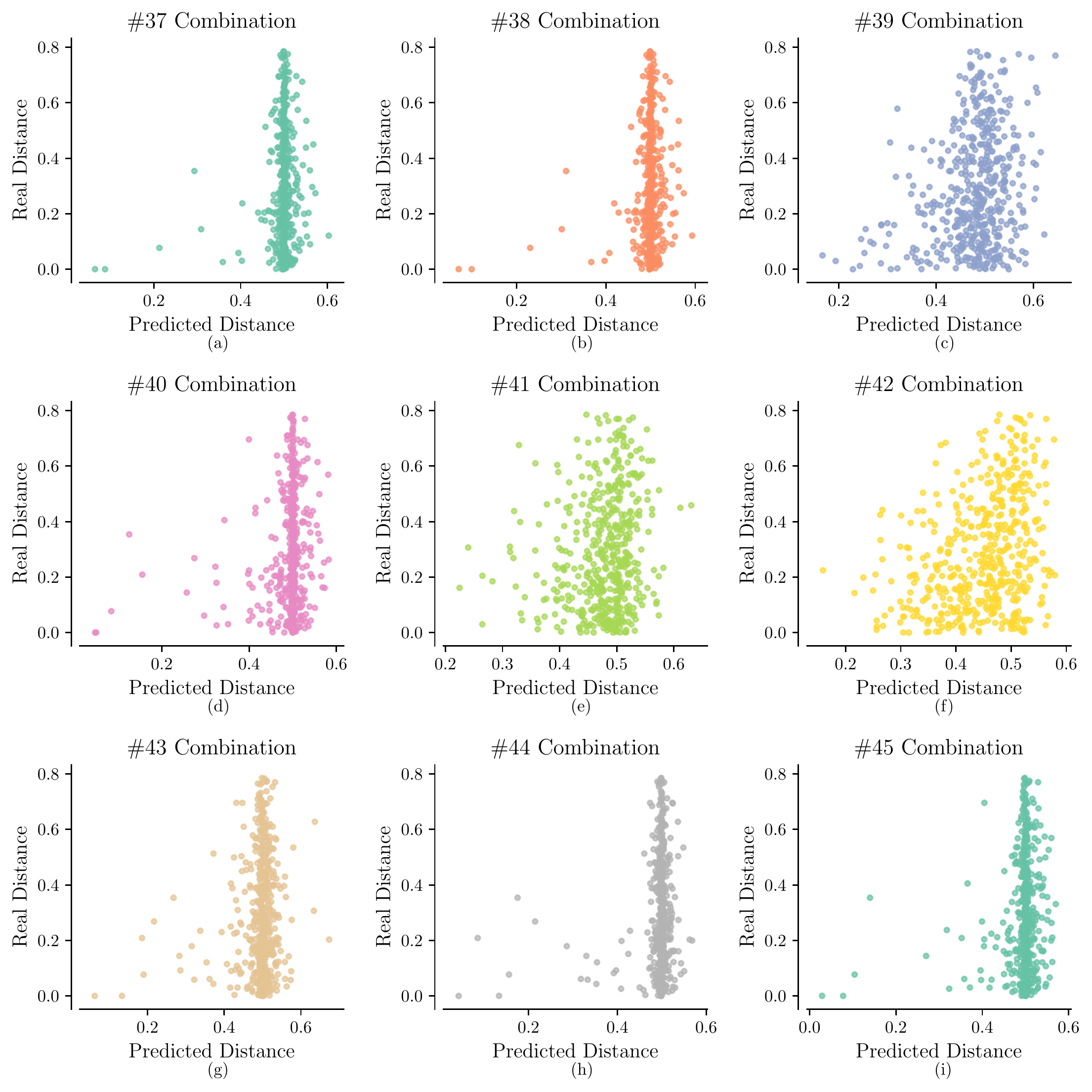}
    \caption{Qualitative performance of different kernel compositions (ranking $\#37\sim\#45$).}
    \label{fig:analysis_4}
\end{figure*}

\clearpage

\begin{table*}[h!]
  \centering
  \begin{tabular}{c|c|c|c|c|c} 
    \toprule
        \textbf{Rank} & \textbf{Figure} & \textbf{Task Kernel} & \textbf{Configuration Kernel} & \textbf{Fidelity Kernel} & \textbf{ELBO} \\
    \midrule
    \midrule
        46 & Figure~\ref{fig:analysis_5} (a) & \texttt{DeepPoly} & \texttt{Flat} & \texttt{Matern} & $0.8844$ \\
        47 & Figure~\ref{fig:analysis_5} (b) & \texttt{DeepLinear} & \texttt{Flat} & \texttt{AccCurve} & $0.4301$ \\
        48 & Figure~\ref{fig:analysis_5} (c) & \texttt{DeepLinear} & \texttt{Flat} & \texttt{RBF} & $0.3416$ \\
        49 & Figure~\ref{fig:analysis_5} (d) & \texttt{DeepPoly} & \texttt{DeepPoly} & \texttt{Matern} & $0.3053$ \\
        50 & Figure~\ref{fig:analysis_5} (e) & \texttt{MTBO} & \texttt{Flat} & \texttt{AccCurve} & $0.2985$ \\
        51 & Figure~\ref{fig:analysis_5} (f) & \texttt{DeepLinear} & \texttt{Flat} & \texttt{Matern} & $0.2822$ \\
        52 & Figure~\ref{fig:analysis_5} (g) & \texttt{MTBO} & \texttt{Flat} & \texttt{RBF} & $0.1433$ \\
        53 & Figure~\ref{fig:analysis_5} (h) & \texttt{MTBO} & \texttt{Flat} & \texttt{Matern} & $0.1406$ \\
        54 & Figure~\ref{fig:analysis_5} (i) & \texttt{DeepPoly} & \texttt{DeepPoly} & \texttt{Fabolas} & $0.1350$ \\
    \bottomrule
  \end{tabular}
  \caption{Quantitative performance of different kernel compositions (ranking $\#46\sim\#54$).}
  \label{tab:analysis_5}
\end{table*}

\begin{figure*}[h!]
    \centering
    \includegraphics[width=0.87\textwidth]{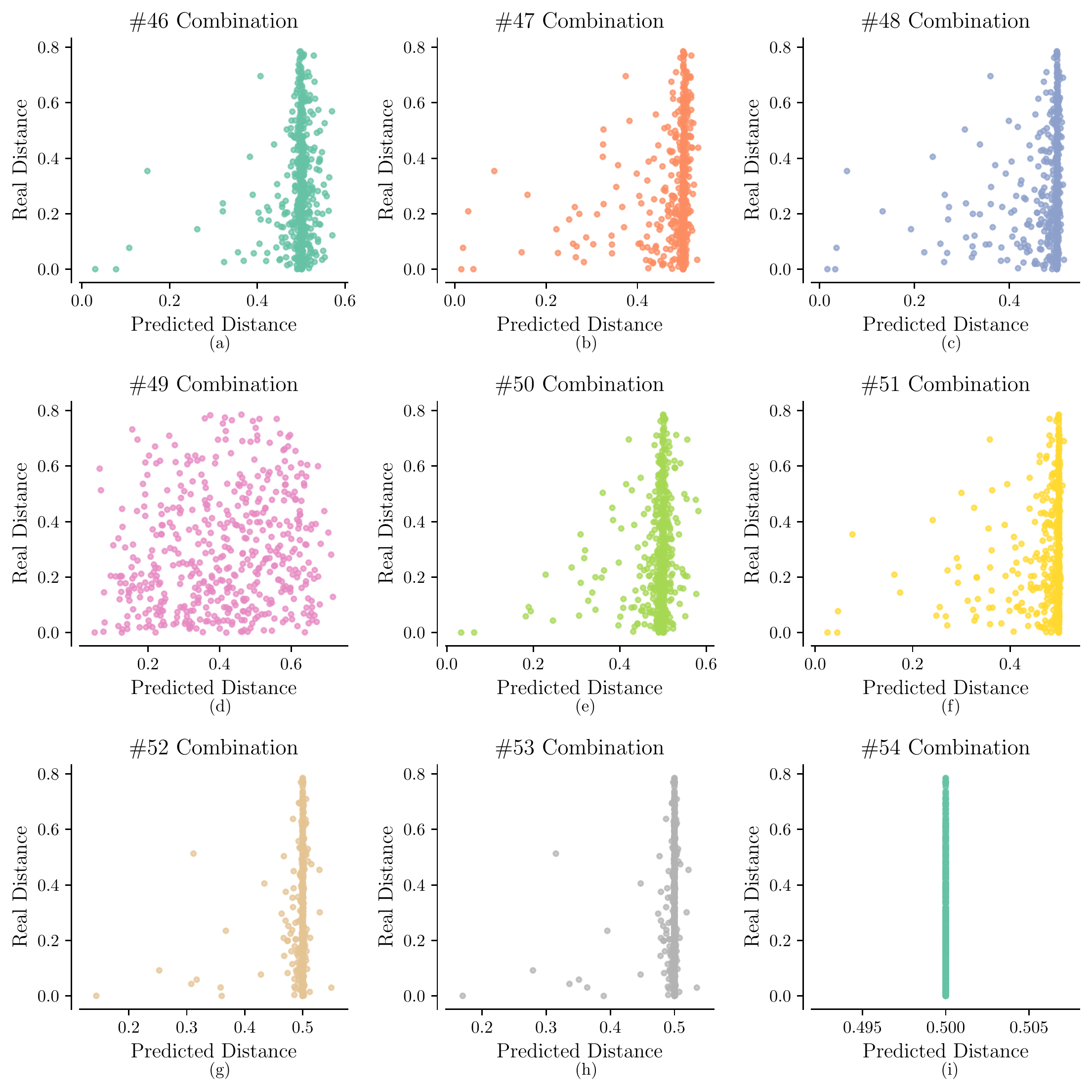}
    \caption{Qualitative performance of different kernel compositions (ranking $\#46\sim\#54$).}
    \label{fig:analysis_5}
\end{figure*}

\clearpage

\begin{table*}[h!]
  \centering
  \begin{tabular}{c|c|c|c|c|c} 
    \toprule
        \textbf{Rank} & \textbf{Figure} & \textbf{Task Kernel} & \textbf{Configuration Kernel} & \textbf{Fidelity Kernel} & \textbf{ELBO} \\
    \midrule
    \midrule
        55 & Figure~\ref{fig:analysis_6} (a) & \texttt{OptiLand} & \texttt{DeepLinear} & \texttt{Fabolas} & $0.1350$ \\
        56 & Figure~\ref{fig:analysis_6} (b) & \texttt{DeepLinear} & \texttt{Flat} & \texttt{Fabolas} & $0.1349$ \\
        57 & Figure~\ref{fig:analysis_6} (c) & \texttt{DeepLinear} & \texttt{Tree} & \texttt{Fabolas} & $0.1349$ \\
        58 & Figure~\ref{fig:analysis_6} (d) & \texttt{DeepLinear} & \texttt{DeepLinear} & \texttt{Fabolas} & $0.1349$ \\
        59 & Figure~\ref{fig:analysis_6} (e) & \texttt{DeepPoly} & \texttt{Flat} & \texttt{Fabolas} & $0.1349$ \\
    \bottomrule
  \end{tabular}
  \caption{Quantitative performance of different kernel compositions (ranking $\#55\sim\#59$).}
  \label{tab:analysis_6}
\end{table*}

\begin{figure*}[h!]
    \centering
    \includegraphics[width=0.87\textwidth]{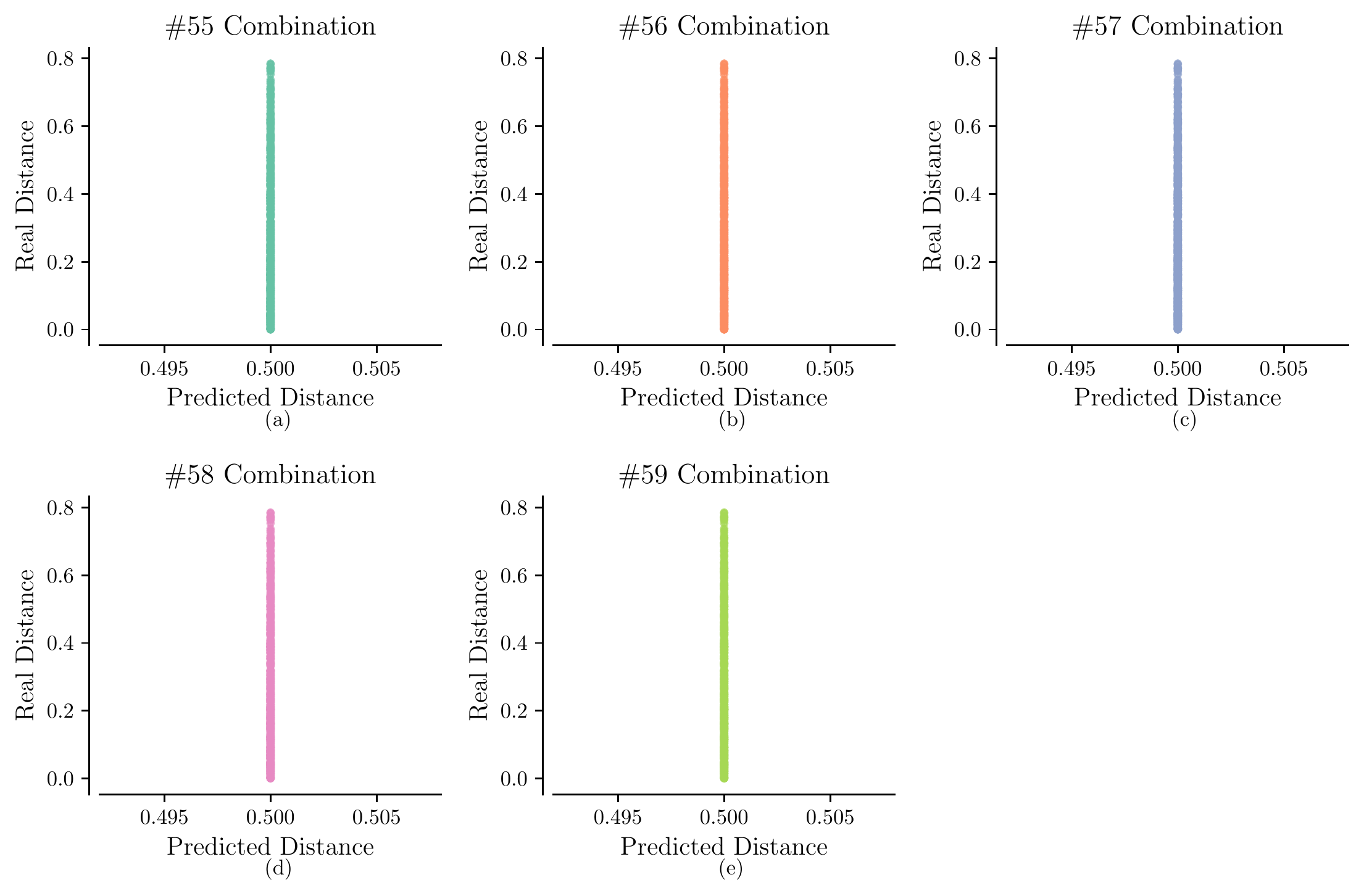}
    \caption{Qualitative performance of different kernel compositions (ranking $\#55\sim\#59$).}
    \label{fig:analysis_6}
\end{figure*}

\clearpage

\begin{table*}[h!]
  \centering
  \begin{tabular}{c|c|c|c|c|c} 
    \toprule
        \textbf{Rank} & \textbf{Figure} & \textbf{Task Kernel} & \textbf{Configuration Kernel} & \textbf{Fidelity Kernel} & \textbf{ELBO} \\
    \midrule
    \midrule
        60 & Figure~\ref{fig:analysis_7} (a) & \texttt{DeepPoly} & \texttt{Tree} & \texttt{Fabolas} & $0.1349$ \\
        61 & Figure~\ref{fig:analysis_7} (b) & \texttt{MTBO} & \texttt{DeepLinear} & \texttt{Fabolas} & $0.1342$ \\
        62 & Figure~\ref{fig:analysis_7} (c) & \texttt{MTBO} & \texttt{Flat} & \texttt{Fabolas} & $0.1342$ \\
        63 & Figure~\ref{fig:analysis_7} (d) & \texttt{MTBO} & \texttt{DeepPoly} & \texttt{Fabolas} & $0.1342$ \\
        64 & Figure~\ref{fig:analysis_7} (e) & \texttt{MTBO} & \texttt{Tree} & \texttt{Fabolas} & $0.1342$ \\
    \bottomrule
  \end{tabular}
  \caption{Quantitative performance of different kernel compositions (ranking $\#60\sim\#64$).}
  \label{tab:analysis_7}
\end{table*}

\begin{figure*}[h!]
    \centering
    \includegraphics[width=0.87\textwidth]{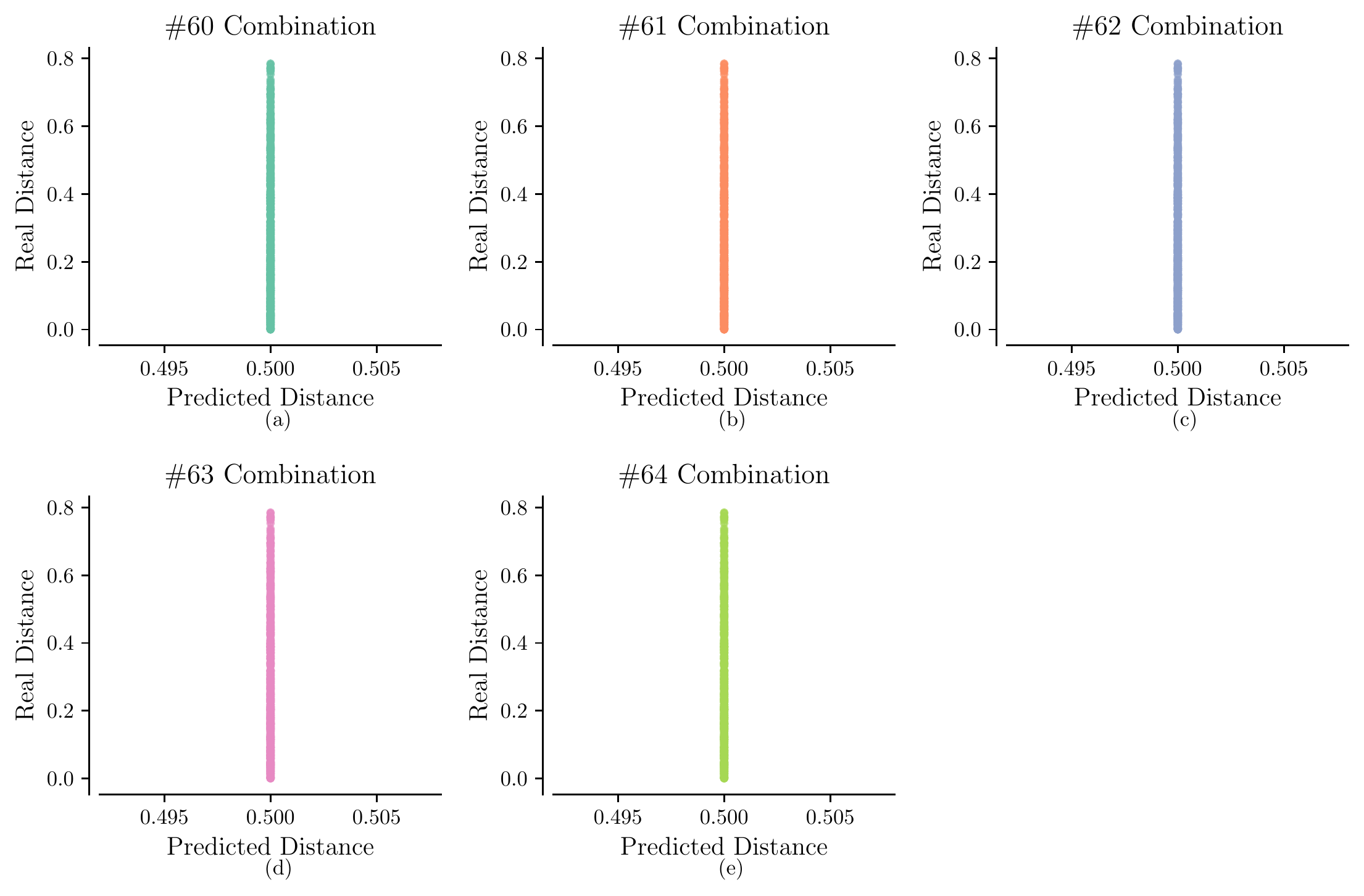}
    \caption{Qualitative performance of different kernel compositions (ranking $\#60\sim\#64$).}
    \label{fig:analysis_7}
\end{figure*}

\clearpage
\section{Experiment Details} \label{app:3}

\begin{figure*}[h!]
    \centering
    \includegraphics[width=0.9\textwidth]{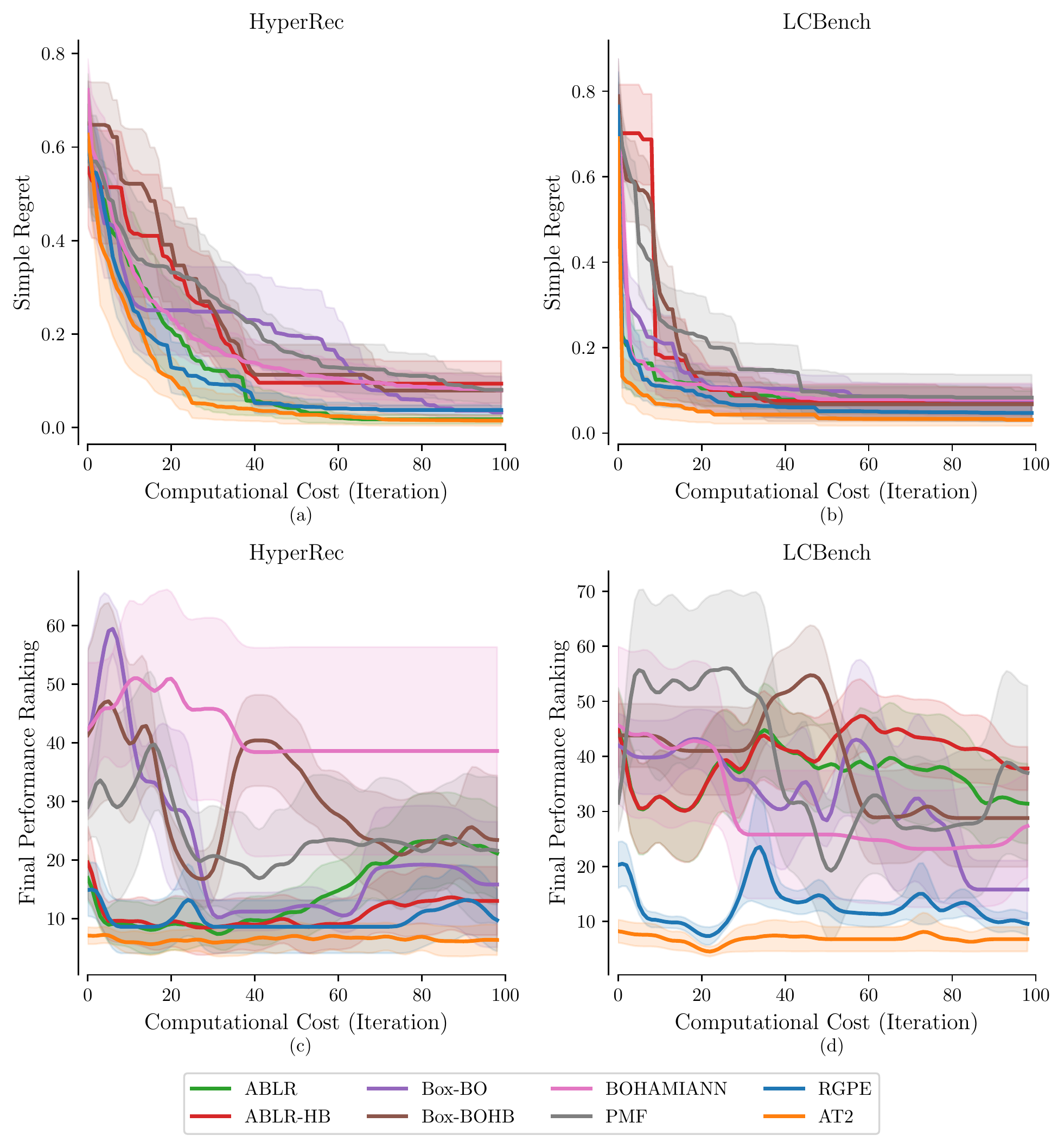}
    \caption{Performance of methods on \textrm{HyperRec} and \textrm{LCBench}. The results based on two metrics (simple regret and final performance ranking) are averaged across five train-test task pairs for each database. Lower is better. The predicted final performance rankings are smoothed with a hamming window of $10$ iterations. The shaded regions represent one standard error of each method. Our proposed \texttt{AT2} method consistently achieves lower simple regrets and final performance rankings.}
    \label{fig:experiment_std}
\end{figure*}

We compare the proposed \texttt{AT2} method against seven hyperparameter transfer learning baselines, based on our offline-computed database \textrm{HyperRec} and another real-world database \textrm{LCBench} \cite{zimmer2020auto}.
The hyperparameters are chosen from a grid search: $\{\numprint{1000}, \numprint{2000}, \numprint{5000}\}$ for the number of inducing points, $\{100, 200, 500\}$ for the number of epochs, $\{0.005, 0.01, 0.02\}$ for the learning rate, and $\{0.25, 0.5, 1\}$ for $\eta$ in the Max-Trial-GP-UCB acquisition function.
Figure~\ref{fig:experiment_std} shows the quantitative performance of \texttt{AT2} and other baselines with one standard error. 
We perform our experiments on an AWS P2 instance with one K80 GPU. It takes around one hour for \texttt{AT2} to finish training and to run $100$ queries on one train-test task pair. 
The detailed experiment setup is explained in Section~\ref{sec:4}.

\end{document}